\documentclass[10pt,journal,compsoc]{IEEEtran}
\usepackage{ragged2e}

\usepackage{ulem}
\usepackage{times}
\usepackage{epsfig}
\usepackage{graphicx}
\usepackage{amsmath}
\usepackage{amssymb}
\usepackage{mathrsfs}
\usepackage{multirow}
\usepackage{makecell}
\usepackage{epstopdf}
\usepackage{amsthm}
\usepackage{bbm}
\usepackage{algorithm}
\usepackage{algorithmic}
\usepackage{color} 

\ifCLASSOPTIONcompsoc
  \usepackage[nocompress]{cite}
\else
  \usepackage{cite}
\fi
\ifCLASSINFOpdf
\else
\fi
\begin{document}
%
\title{T2TD: Text-3D Generation Model based on Prior Knowledge Guidance}
%
%
%
%

\author{Weizhi Nie,~
        Ruidong Chen,~
        Weijie Wang$^*$,~
        Bruno Lepri,~
        Nicu Sebe,~\IEEEmembership{Senior Member,~IEEE}
\IEEEcompsocitemizethanks{
\IEEEcompsocthanksitem Weizhi Nie and Ruidong Chen are with the school of electrical and information engineering, Tianjin University. E-mail: \{weizhinie, chenruidong
\}@tju.edu.cn.
\IEEEcompsocthanksitem Weijie Wang and Nicu Sebe are with the Department of Information Engineering and Computer Science, University of Trento, Trento 38123, Italy. E-mail:\{weijie.wang, niculae.sebe\}@unitn.it. $^*$Corresponding author.
\IEEEcompsocthanksitem Bruno Lepri is with Fondazione Bruno Kessler, Italy. E-mail: lepri@fbk.eu
}
\thanks{Manuscript received April 19, 2005; revised August 26, 2015.}

}

%
%

\markboth{Journal of \LaTeX\ Class Files,~Vol.~14, No.~8, August~2015}%
{Shell \MakeLowercase{\textit{et al.}}: Bare Demo of IEEEtran.cls for Computer Society Journals}
%



\IEEEtitleabstractindextext{%
\begin{abstract}
In recent years, 3D models have been utilized in many applications, such as auto-driver, 3D reconstruction, VR, and AR. However, the scarcity of 3D model data does not meet its practical demands. Thus, generating high-quality 3D models efficiently from textual descriptions is a promising but challenging way to solve this problem. 
In this paper, inspired by the ability of human beings to complement visual information details from ambiguous descriptions based on their own experience, we propose a novel text-3D generation model (T2TD), which introduces the related shapes or textual information as the prior knowledge to improve the performance of the 3D generation model.
In this process, we first introduce the text-3D knowledge graph to save the relationship between 3D models and textual semantic information, which can provide the related shapes to guide the target 3D model generation.
Second, we integrate an effective causal inference model to select useful feature information from these related shapes, which removes the unrelated shape information and only maintains feature information that is strongly relevant to the textual description. 
Meanwhile, to effectively integrate multi-modal prior knowledge into textual information, we adopt a novel multi-layer transformer structure to progressively fuse related shape and textual information, which can effectively compensate for the lack of structural information in the text and enhance the final performance of the 3D generation model.
The final experimental results demonstrate that our approach significantly improves 3D model generation quality and outperforms the SOTA methods on the text2shape datasets. 
\end{abstract}

}

\maketitle

\IEEEdisplaynontitleabstractindextext

%
\IEEEpeerreviewmaketitle

\section{Introduction}
\IEEEPARstart{I}{n} recent years, 3D models have been applied to many applications, such as fabrication, augmented reality, and education. An increasing number of researchers have focused on how to satisfy the huge industrial demands for 3D models. Obtaining 3D models via professional software (such as Maya, Blender, and 3DSMAX) is a laborious manual process that requires specific expertise for the user. Thus, obtaining 3D models more efficiently and concisely has become a hot topic recently. However, the complex visual and structural information of the 3D models creates substantial challenges and difficulties.
Consequently, different types of approaches have been proposed to handle this problem \cite{3Dgen_3Dvaegan,3Dgen_grass,3Dgen_spgan,3dgen-substructure}, and several works have attempted to recover 3D information from 2D images (rendered view\cite{3Drec_3DR2N2,3Drec_pix2vox,3Drec_pix2vox++,3Drec_pami,pix2mesh++}, scene \cite{3Dscene_pami,3Dscene_m2,3Dscene_m3}, sketch\cite{3Drec_sketch,3Drec_sketch2,3Drec_sketch3}). In addition, some cross-modal 3D retrieval methods \cite{nie2019hgan,3drev1,3drev2} are used to search and match the 3D models in databases, which reduces the difficulty of acquiring models, but still falls short of human expectations in terms of the accuracy and matching requirements.

\begin{figure}[t]
	\centering
	\includegraphics[width=1\linewidth]{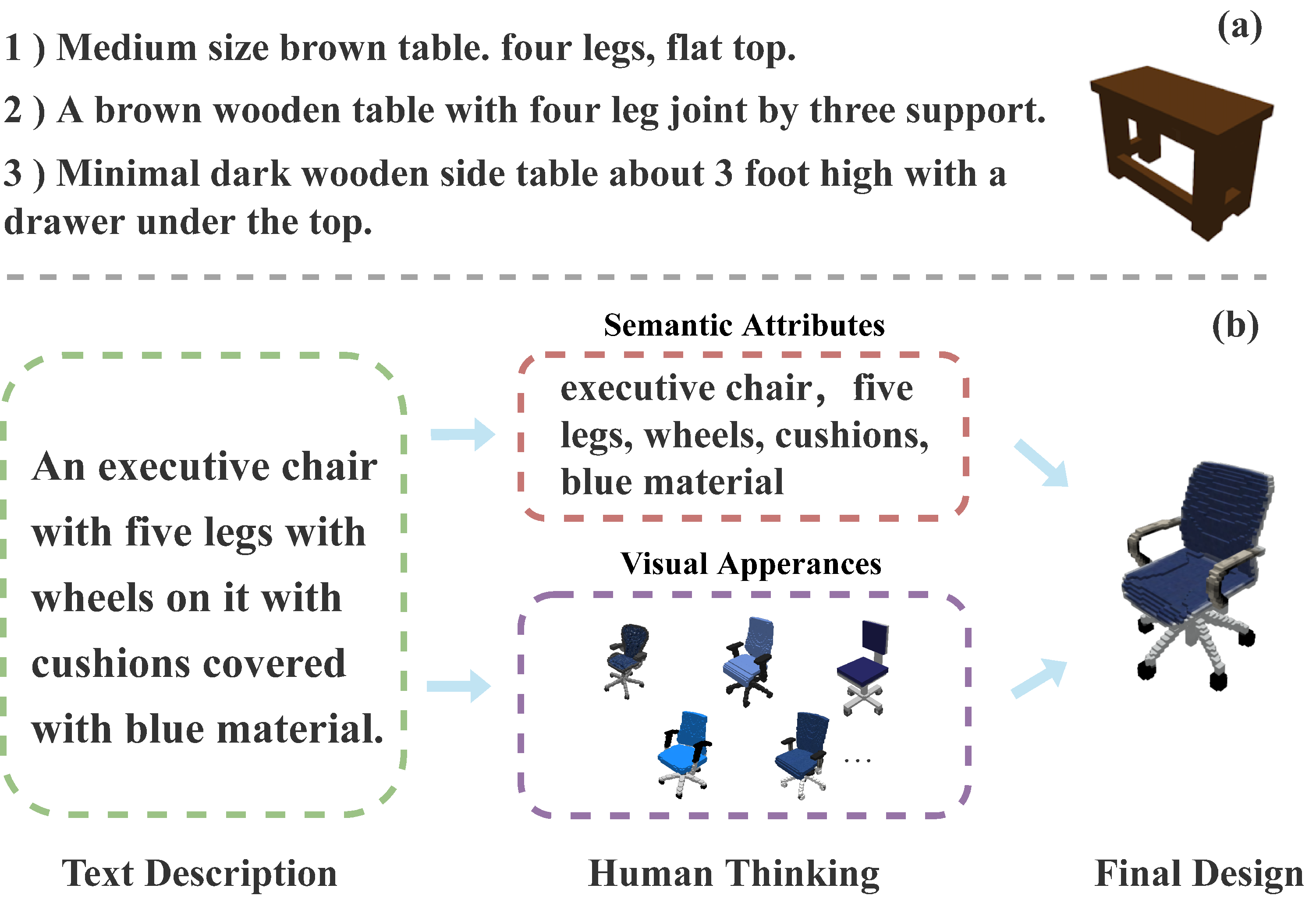}
  \caption{a) A single caption can only describe part of the appearance of a 3D object, and ambiguous descriptions may cause difficulties for text-3D works. b) Inspired by the human thinking mode, we think the two types of prior knowledge (semantic attributes and related shapes) can be used to provide more detailed information and enhance the text-3D generation task.}
  \label{fig1}
\end{figure}

A more convenient way of acquiring 3D models is to use natural language. Based on natural language, humans just need to express their thoughts precisely without the need to provide any additional information, such as images or similar 3D objects. However, this way does not meet well with the expectation of humans for 3D models. In recent years, only a few works have focused on this challenging field. Text2Shape\cite{text2shape} is the first work to generate colored 3D shapes from flexible natural language. Their work uses a similar idea with several cross-modal representation methods \cite{reed1,reed2} and consists of two parts. First, they learn the joint representations of text and 3D shapes. Then, they use the learned text embedding as input conditions to predict the corresponding 3D shapes directly by training a GAN structure. However, the method only generates an approximate appearance that matches the input text and does not achieve a sufficiently satisfactory generation quality.

The recent work\cite{text2shapecvpr} adopts a more straightforward approach to guide the 3D model generation using textual information, which first trains a 3D autoencoder(AE) and directly projects the text features into the learned 3D feature space. Using the aligned text feature to feed with the learned 3D shape decoder, their methods can achieve a favorable 3D shape generation performance.

However, due to the huge cross-modal gap between text and 3D model, the aforementioned methods still have limitations when faced with some specific situations:
\begin{itemize}
  \item Rough description: A single sentence cannot fully evolve all the geometric information. Meanwhile, many sentences may also lack detailed descriptions, especially a 3D structure information description. We need to consider how to supplement this information.
  \item Diversity description: Different people often have different descriptions of the same object. The flexibility of natural language also causes ambiguities in learning stable cross-modal representations. The lack of large-scale text-3D datasets further amplifies this kind of ambiguity and leads to the uneven quality of the generated 3D shapes. 
\end{itemize}

\subsection{Motivation}
In light of our analysis, we hope the 3D generation model can automatically introduce some prior knowledge in the same way as humans. Fig.\ref{fig1} (b) shows this motivation. When we say: ``an executive chair with five legs with wheels on it with cushions covered with blue material''. A human can think about characteristics such as ``five legs'', ``wheels'', and related 3D models. These pieces of information can help humans synthesize the final 3D model to handle the rough description problem. Inspired by this, we hope to leverage similar prior knowledge to assist with text-3D generation methods. In this process, the diversity prior knowledge can help us to synthesize diversity 3D models and handle the 
diversity description problem. Specifically, we need to address the following fundamental challenges:
\begin{itemize}
  \item How to define the format of prior knowledge, which maintains the latent geometric structure information and the related 3D models in a human-like way. We also need to ensure that this prior knowledge is useful for improving the model generation;
  \item How can prior knowledge be achieved based on the input textual information? We need to capture the correlation between the prior knowledge and the input text. This correlation can also be used to search the related prior knowledge based on the text in the testing step;  
  \item How to design the generation network to leverage the prior knowledge to enhance the geometric detail and improve the generation qualities.
\end{itemize}
  
In this paper, we propose a novel 3D generation model via textual information (T2TD) address these issues. Specifically, our framework is built upon the existing text-3D data set\cite{text2shape}, which explicitly define the entity and edge to construct a text-3D knowledge that maintains the correlation between the text and 3D shape, as well as the related 3D shape and attributes.
Here, we define the related 3D shape and textual attributes as prior knowledge. The knowledge graph can save the prior knowledge and introduce more knowledge information as the data increase.
In the generation step, we apply \cite{xiong2021knowledge} to search the prior knowledge from the knowledge graph according to the text description. However, it should be noted that the searched shapes' prior knowledge is only similar to the text description, but not completely consistent. To remove irrelevant shape information, we propose an effective casual model to select shape information from the prior shape knowledge, selecting feature information strongly related to the text description. 
Finally, we apply a multi-layer transformer structure to progressively fuse the prior knowledge and the textual attribute information, which compensates for the lack of structural information in the text and enhances the final performance of the 3D generation models. Compared with a traditional generation model, we add prior knowledge into the generation network, which can improve the final generation performance. The final experimental results also demonstrate that our approach significantly improves the 3D model generation quality and performs favorably against the SOTA methods on the Text2Shape\cite{text2shape} datasets.

\subsection{Contribution}
The contributions of this paper can be summarized as follows:
\begin{itemize}
\item We define the format of prior knowledge and first propose a novel 3D shape knowledge graph to bridge the gap between the text and the 3D models. In addition, using our constructed 3D commonsense knowledge graph, we can save and achieve richer prior knowledge;
\item We propose a novel casual inference model to select the related feature and remove the unrelated 3D information from the prior knowledge, which can achieve more useful information for final 3D model generation; 
\item We propose a novel text-3D generation model (T2TD), which can fuse the useful prior knowledge and generate the related 3D model according to the textual information and greatly reduces the difficulty of obtaining 3D model data;
\end{itemize}

The remainder of this article is organized as follows. Section 2 presents several related works. Section 3 provides the details of our approach. The corresponding experimental results and analysis are given in Section 4. Finally, we discuss the limitations and our future work and conclude this paper in Section 5.

\begin{figure*}[t]
	\centering
	\includegraphics[width=1\linewidth]{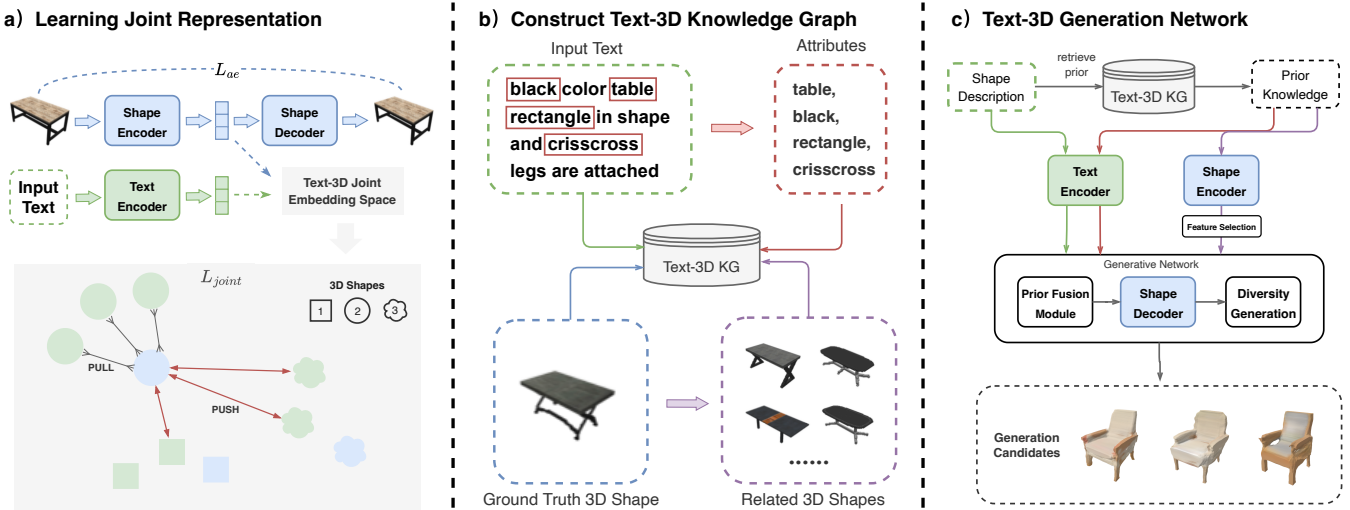}
  \caption{The overall framework of T2TD mainly includes three parts: a) A pre-trained representation module, which learns the 3D geometric information through an autoencoder and learns text-3D joint representations through cross-modal contrastive learning. b) Constructing the text-3D knowledge graph to structurally associate the texts and 3D shapes, which is used to provide prior information for the generative network. c) A text-3D generation network to leverage text input and retrieve prior knowledge to generate 3D shapes.}  
  \label{framework}
\end{figure*}

\section{Related Works}
\subsection{3D Shape Generation}
Recently, there has been a considerable amount of work devoted to the task of 3D shape generation. In the traditional methods, the frameworks always generate 3D data for a specific 3D shape representations, such as 3D voxels\cite{3Dgen_3Dvaegan,3Dgen_grass,3Drec_3DR2N2,3Drec_pix2vox,3Drec_pix2vox++}, point clouds\cite{3Dgen_treegan,3Dgen_spgan,3Dgen_diffusion,3Dgen_diffusion2} and meshes\cite{pix2mesh,pix2mesh++,pix2mesh_pami}. However, these methods have a common limitation is that the generated 3D shapes are limited in a specific resolution, which causes inflexibility in practical applications.

To solve the problem, recent works start to explore the implicit functions\cite{implicit-im-net,implicit-deepsdf,implicit-disn,implicit-peoplesdf} to represent 3D shapes. The implicit function-based methods calculate the 3D model surface by encoding the point coordinates and predicting occupancy of each position, together with the Marching Cubes algorithm, which can generate 3D shapes with arbitrary resolution. In addition to the 3D generation task\cite{implicit-im-net,implicit-autosdf,text2shapecvpr}, the implicit functions have been used in many other tasks, such as image-based 3D reconstruction\cite{implicit-occnet,implicit-autosdf} and 3D shape deformation tasks\cite{implicit-template,implicit-deformation}. 


\subsection{Text-Image Generation}
With the publications of large-scale text-image datasets\cite{flower,COCO,yfcc100m}, remarkable progress has been made in the field of text-image representations\cite{reed1,reed2}. Many related works begin to focus on how to use natural language to get high-quality and semantically consistent images. In the early research of this field, many approaches\cite{gan-int-cls,stackgan,stackgan++} leverage the GAN\cite{GAN} structure by feeding text embeddings as the conditional input to generate corresponding images through natural language. And the subsequent works\cite{AttnGAN, SEGAN, MirrorGAN,DMGAN, RiFeGAN,df-gan} improved the GAN-based framework from different aspects. Recently, several approaches have been proposed \cite{Dalle, CogView} which are not based on GAN structure and get favorable generation performance.

Compared with 2D images, the 3D shape expresses the complete spatial structure of a real object and has rich geometric information. As for the text-3D generation task, the lack of large-scale text-3D datasets also poses difficulties to the research of this kind of task. Therefore, we hope to use the knowledge graph to make full use of the existing dataset and improve the text-3D performance.

\subsection{Text-3D Generation}
Currently, most of the related text-3D work is engaged to handle the text-3D retrieval\cite{text2shape,shapecaptioner,y2seq2seq} or 3D shape captioning\cite{shapecaptioner,y2seq2seq} tasks, there are only a few works engaged in addressing the challenging task of using natural language to generate 3D shapes. Text2shape\cite{text2shape} adopts a similar idea with text-to-image generation methods\cite{gan-int-cls} to train the generator with a GAN\cite{GAN} structure and directly predict 3D volume as its output. 
However, due to the inadequate joint learning of natural language and 3D shapes, it fails to generate favorable 3D shapes consistent with the input text. 

Text-Guided\cite{text2shapecvpr} takes an alternative approach to solve this problem. It aligns text and shape features in the same embedding space learned by the 3D autoencoder. As a result, the extracted text features are directly used to generate 3D models using the 3D decoder. In addition, to diversify the generation results, they adopt an IMLE-based (Implicit Maximum Likelihood Estimation) generator to apply random noise on the learned text feature, which avoids the mode collapse of GANs.

There are also some other works to achieve the task of text-3D generation from different perspectives. Such as \cite{text2mesh} engage in generating high-quality textures for 3D mesh according to text descriptions, and \cite{clip-forge} exploit the CLIP\cite{clip} model to generate approximate shape from the description in a Zero-Shot way. Different from the manner to generate 3D models of them, the aim of our method is to use text information to directly generate 3D shapes with semantic consistent details of structure and color information, we do not treat them as competitors.

\subsection{Visual Generation via Prior Knowledge}
Several previous works have successfully introduced prior knowledge into cross-modal visual generation tasks. To overcome the deficiency of detailed 2D information in the text descriptions, the RifeGAN\cite{RiFeGAN} utilizes an external knowledge database to retrieve similar sentences according to the input text descriptions to supply detailed semantic information. In the 3D field, Mem3D\cite{3Drec_mem} utilizes retrieved 3D shapes to serve as the prior knowledge to help the network recover 3D information from 2D images with complex backgrounds and heavy occlusions.

In our proposed method, we use both of the above types of prior knowledge to assist in the text-3D generation tasks, which are from semantic and visual two perspectives. With the corresponding generative networks, we can effectively integrate prior knowledge into the generation process.

\section{Approach}
In this section, we detail our approach. Fig.\ref{framework} shows the framework which includes three key parts. 1) Pre-trained representation model: it is used to learn the textual and 3D model features in the common feature space. The aim of this operation is to build the correlation between the text and the 3D shape for the knowledge graph construction; 2) 3D shape knowledge graph: we define the entity, edge, and related attribute information to save the prior knowledge in the knowledge graph, which can be used to search and associate the related shapes and semantic information based on query text; 3) 3D model generation module: it is used to fuse the cross-model prior knowledge information to make up for the lack of structural information, and generate the target 3D model. We will detail these modules in the next subsections.

\subsection{Pre-trained Representation Module}
This module is concerned with data preprocessing. Specifically, we exploit a 3D shape autoencoder to fully learn the representations of the 3D shapes with rich geometric and color information. In addition, we propose a joint feature representation model to train the text in the same latent space with the 3D shapes. We will detail these modules in the following subsections.  
\begin{figure}[t]
	\centering
	\includegraphics[width=1.0\linewidth]{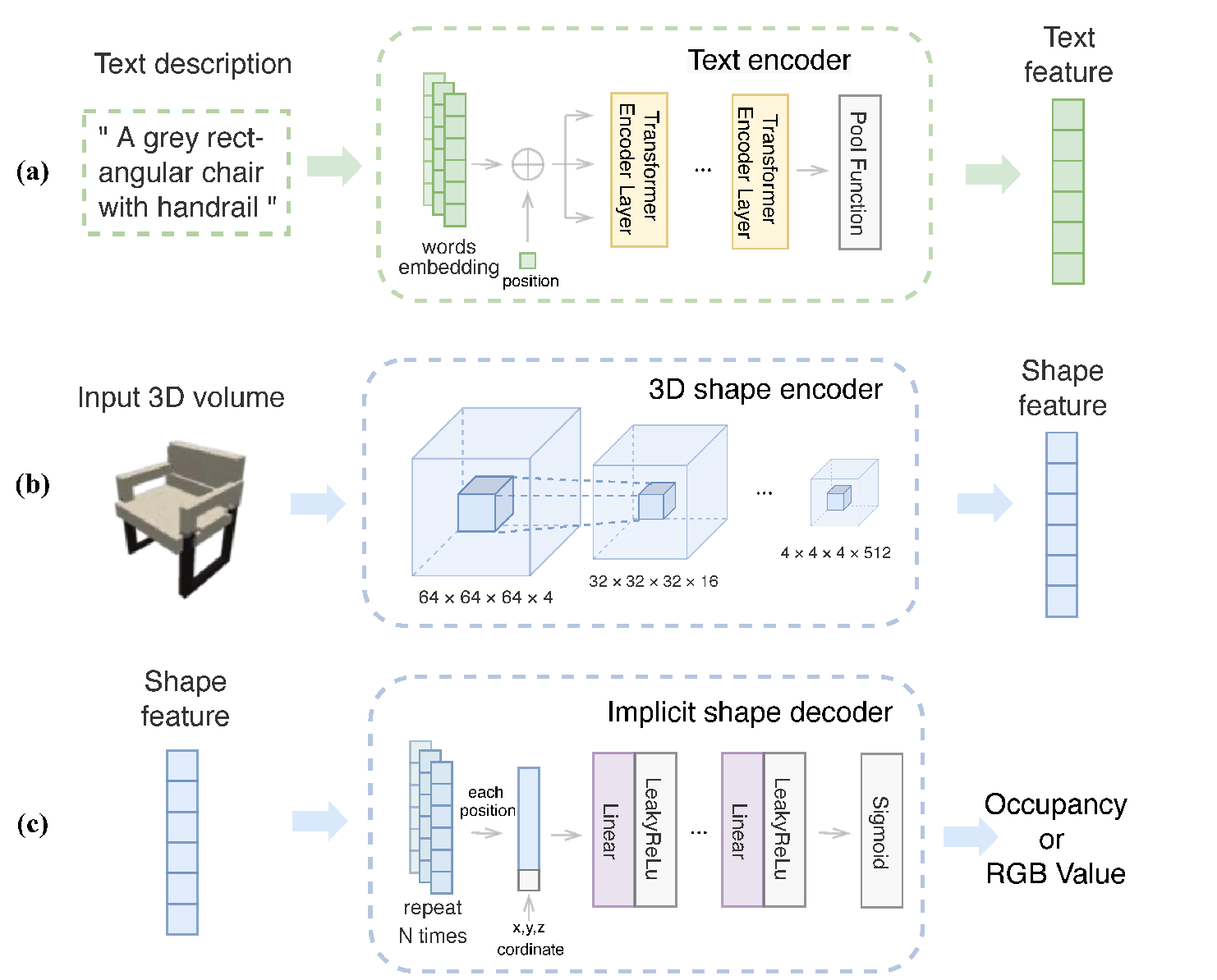}
  \caption{The basic architecture of the encoder networks. (a) The transformer-based text encoder, it converts the input text description into a global sentence feature. (b)The CNN-based 3D shape encoder, it converts the colored 3D volume into a global 3D feature. (c)The implicit shape decoder, it takes a 3D shape feature with a  point coordinate as input and predicts the occupancy probability or the RGB value of each sampled position. }
  \label{encoder}
\end{figure}

\subsubsection{Text Encoder}
The text encoder $E_t$ is a 6-layer transformer encoder\cite{transformer}, which is shown in Fig. \ref{encoder}. Here, the structure of the transformer can effectively improve the performance of textual embeddings, which have been proven by many classic approaches\cite{bert, clip}. We first extract the embeddings $x_t\in \mathbb{R}^{L\times{e_w}}$ of the query text, where $L$ is the length of the sentence and $e_w$ indicates word embeddings. Then, $E_t$ receives $x_t$, Here, the transformer encoder consists of the multi-head self-attention layers, which attempt to find more latent correlation information between the words and reduce the redundant information to improve the performance of the final textual representation. The transformer output is operated by the pool function and achieves the final text feature $f_t \in \mathbb{R}^{d}$.

\subsubsection{Shape Encoder}
We use the 3D volume as the input to learn the information from 3D models. The basic structure of the networks is shown in Fig. \ref{encoder}. Inspired by the basic method in previous work, our voxel encoder $E_v$ consists of $5$ 3D convolutional blocks to take a 3D input $x_v\in \mathbb{R}^{{r_v}\times{r_v}\times{r_v}\times4}$ and calculate it to the 3D shape features $f_s \in \mathbb{R}^{d}$, where $r_v$ represents the resolution of the input 3D shapes, and $d$ represents the dimension of the extracted features.

\subsubsection{Implicit Shape Decoders}
Inspired by\cite{implicit-im-net}, we exploit the implicit 3D shape representation as the prediction output of the shape encoder. Here, we sample the 3D volume as an RGBA point sequence $S\in\mathbb{R}^{N\times(1+3)}$, with a sampled sequence representing the 3D spatial position of each point, where $N$ represents the number of sampled points. Respectively, we applied shape decoder $D_s$ and color decoder $D_c$ to predict shape occupancy and RGB color for each point. By concatenating the point position $p$ with the extracted $f_s$, $D_s$ predicts the shape occupancy value with five fully-connected and leaky-ReLU layers. $D_c$ has the same architecture as $D_s$ and outputs the predicted RGB color values according to the same point position $p$.

\begin{figure*}[ht] 
	\centering
	\includegraphics[width=1\linewidth]{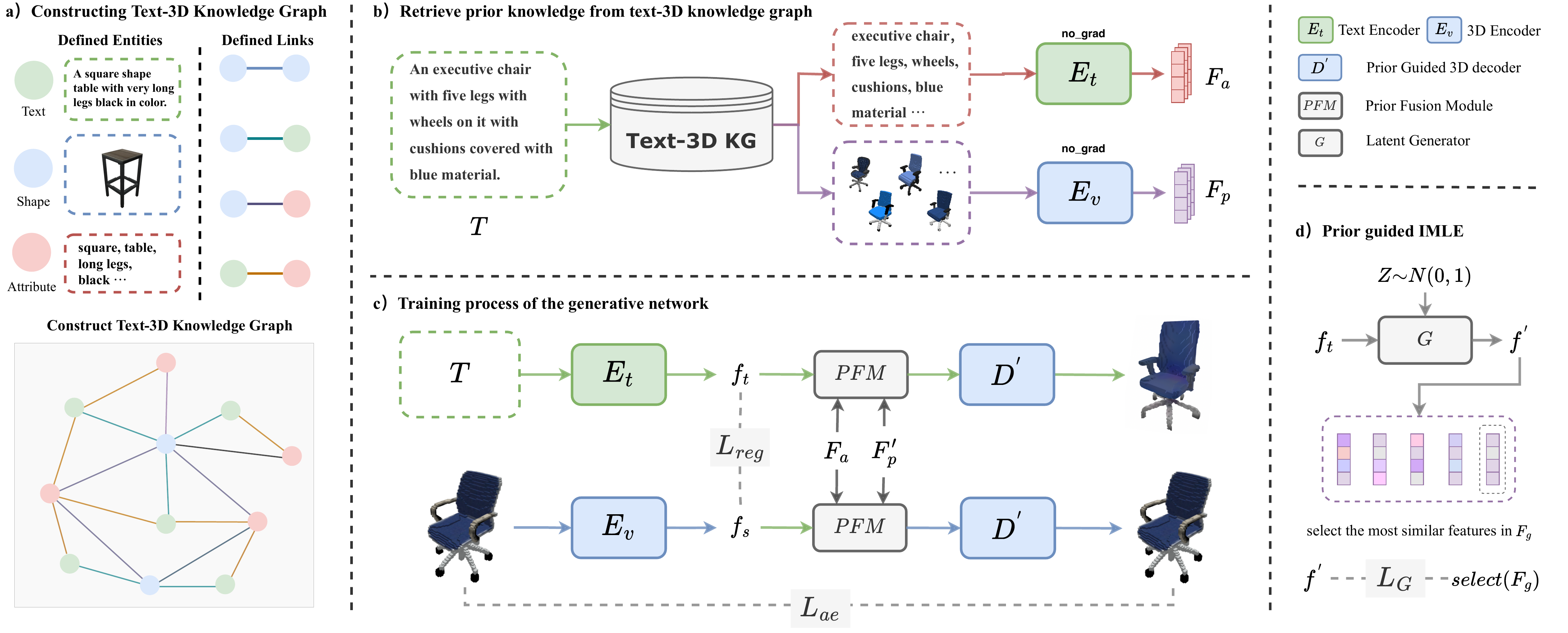}
  \caption{Overview of the framework implementation, which mainly consists of four parts: a) Constructing the knowledge graph by defining the entities and relations in the graph. b)Retrieve two types of prior knowledge and the extracted features by the proposed encoders. c)The training process of the text-3D generative network, which mainly aims to reduce the gap between text and 3D modalities by introducing prior knowledge. d)To further diversify the generation results, adapted to our methods, we propose a prior guided IMLE to fully utilize the prior knowledge.}
  \label{generation}
\end{figure*} 

\subsubsection{Optimization}
To pre-establish the basic relationship between the text and shape information, inspired by ConVIRT\cite{convirt}, we introduce a cross-modal contrastive loss to optimize the pre-trained modules. In a mini-batch with $n$ shape-text pairs, the $i_{th}$ pairs can be represented as $(x_{t_i},x_{v_i})$, which can be defined as the positive pairs. In contrast, the negative pair can be defined as $(x_{t_i},x_{v_j})$ or $(x_{v_i},x_{t_j}),i \neq j$. The loss function can be written as:
\begin{equation}
    \begin{aligned}
      l_{i}^{\mathrm{t} \rightarrow \mathrm{v}} &=-\log \frac{\exp \left(\left\langle E_t(x_{t_i}),E_v(x_{v_i})\right\rangle \right)}{\sum_{j=1}^{n} \exp \left(\left\langle E_t(x_{t_i}),E_v(x_{v_j})\right\rangle \right)},
    \end{aligned}
\end{equation}
\begin{equation}
    \begin{aligned}
      l_{i}^{\mathrm{v} \rightarrow \mathrm{t}} &=-\log \frac{\exp \left(\left\langle E_v(x_{v_i}),E_t(x_{t_i})\right\rangle \right)}{\sum_{j=1}^{n} \exp \left (\left\langle E_v(x_{v_i}),E_t(x_{t_j})\right\rangle \right)},
    \end{aligned}
\end{equation}
where $\left\langle\, \right\rangle$ is the cosine similarity between two feature vectors. We maximize the feature similarity between the positive pairs and minimize the negative pairs. The final cross-modal contrastive loss can be written as:
\begin{equation}
L_{joint}=\frac{1}{n} \sum_{i=1}^{n}\left(\alpha l_{i}^{\mathrm{t} \rightarrow \mathrm{v}}+(1-\alpha) l_{i}^{\mathrm{v} \rightarrow \mathrm{t}}\right),
\end{equation}
where $\alpha \in [0,1]$ is the weight parameter that controls the balance of the loss function between two calculating directions. The introduction of the optimized target can pre-establish the relationship between text and shape information. The learned cross-modal correlation will be further exploited in the knowledge graph's construction step.

In addition, the 3D shape autoencoder architecture ($E_v$,$D_s$,$D_c$) is trained to obtain the geometric and color information for reconstructing the final 3D shape. With the 3D shape feature $f_s$ extracted by $E_v$, $D_s$ and $D_c$ are optimized with :
\begin{equation}
  \begin{aligned}
    L_{ae} = ||D_{s}(f_s\bigoplus p) - I_s||_2 + ||D_{c}(f_s\bigoplus p)\times I_s - I_c||_2 ,
  \end{aligned}
\end{equation}
where the $I_s$ and $I_c$ are the sampled ground truth values of the point occupancy and the color corresponding to the same point position $p$. Here, $D_s$ and $D_c$ are trained to predict the shape and color separately, and the loss function is applied to minimize the L2 distance between the predicted values and the ground truth. To predict the color values according to the point occupancy, the optimization of the predicted color only takes effect on point positions where the occupancy is 1 in the $I_s$.


\subsection{Text-3D Knowledge Graph}
In this work, we propose a novel 3D knowledge graph to save the 3D shape prior knowledge, which can store the association between the natural language and the 3D shapes from multiple perspectives. In the process of knowledge graph construction, we define different entities and relations to map the entire text-3D dataset into a knowledge graph $K$. 
\begin{itemize}
\item \textbf{3D Shape Entity (S):} It represents each 3D shape from the dataset. Here, we utilize the pre-trained 3D shape encoder $E_v$ to extract features of each 3D shape as the shape entity descriptor in $K$.
\item \textbf{Text Entity (T):} The text description of each 3D shape. We extract text features with the pre-trained text encoder $E_t$ as the text entity descriptor.
\item \textbf{Attribute Entity (A):} It can be seen as the sparse semantic label describing the certain perspective of the 3D model. For example, one  3D shape description ``comfortable red color chair with four legs'' has attributes of \{`comfortable', `red', `chair', `four legs'\}. In the proposed framework, we use the keyphrase toolkit\cite{pke} to extract the attribute entities from each 3D shape description. After a manual adjustment, $377$ words and $1,679$ noun phrases and descriptive phrases are finally selected as attribute entities. Similarly, we utilize the pre-trained text encoder $E_t$ to extract features for each selected attribute as entity descriptors.
\end{itemize}

According to these entities, we further define the following relations, which can also be regarded as the edges in the graph:
\begin{itemize}
\item \textbf{Similar Shape Edge (S-S):} It describes the correlation among the 3D model entities. To construct prior relationships, for each 3D shape with its multiple text descriptions, we conduct multiple text-shape retrievals and one shape-shape retrieval using the pre-trained encoders $E_t$ and $E_v$ based on cosine distance. For each 3D shape, we gather all the retrieval results and calculate the similarity scores with their retrieved frequencies and cosine distances. The top $k$ 3D shapes with higher similarity scores are selected to build S-S relations, and each similarity score is set as the weight of the edges;
\item \textbf{Caption Edge (T-S)}: It stores the original correlation between the text and the 3D shapes, and the T-S edge simply links the text entities with its 3D shape. In the application scenario of this knowledge graph, a 3D shape is described by multiple texts. Therefore, in this knowledge graph, a shape entity is often linked by multiple text entities, while a text entity is linked by only one shape entity;
\item \textbf{Attribute Edge (S-A and T-A)}: The T-A edge links text entities and their contained attribute entities, and the S-A edge links the 3D shapes with all its matched attribute entities to their text descriptions. These edges can be used to bridge the relationship between two shape entities or text entities.
\end{itemize}

Based on these definitions, the 3D shape knowledge graph can effectively save clear shape information, attribute information, and textual description information. The different edges can help us to find related textual and shape information according to the query text. 

In general, we are inspired by the mechanism of human thoughts to consider similar shapes (S-S) and attributes (S-A) from two different prior knowledge perspectives. Here, the S-S edge helps us find similar 3D models via the query text. The S-A edge helps us to find the related attribute information. For example, when we obtain the description of an object: ``a red chair has four legs and a threaded backrest''. We can extract the related attribute information: four legs, red, chair, threaded backrest. This attribute information can be utilized to find the related shape information as the shape prior knowledge. 

The mathematical method is described as follows. For a query text $T$, we first find its related attribute entities in the constructed knowledge graph. Then, we apply the text encoder to extract $f_t(T)$ and $f_t(a_i)$ as the feature of the text and attributes respectively. Finally, the multi-entity search method \cite{xiong2021knowledge} is used to search related shape entities as the prior knowledge.
For details, please refer to Algorithm.\ref{algorithms}.
\begin{algorithm}[h]
	\caption{Process of prior knowledge retrieval}
    \label{priorretrieval}
	\begin{algorithmic}[1]
    \REQUIRE 
    text description $T$, text-3d knowledge graph $K$ with entities $\{A,S,T\}$ and edges $\{S-S,T-S,T-A,S-A\}$
    \ENSURE related shapes $P_s=\{s_1,s_2,...s_m\}$ and related attributes $P_a=\{a_1,a_2,...a_n\}$
    \STATE Match existing attribute entities with $T$,
    \FOR{$a_i$ in $A$}
      \STATE \textbf{If} $a_i$ in $T$ \textbf{then} insert $a_i$ into $P_a$
    \ENDFOR
    \STATE Search mode $(\ P_a\ ,\ A-S\ ,\ ?\ )$ in $K$. Get $P_s^{'}$
    \STATE Search mode  $(\ P_s^{'}\ ,\ S-S\ ,\ ?\ )$ in $K$. Get $P_s^{''}$
    \STATE Set $P_s$ with top $m$ retrieved 3D object of $\{s_1,s_2,...s_m\}$ sorted by weight scores from $P_s^{'}$ and $P_s^{''}$.
    \STATE \textbf{return} $P_s=\{s_1,s_2,...s_m\}$,$P_a=\{a_1,a_2,...a_n\}$
	\end{algorithmic}
	\label{algorithms}
\end{algorithm}

\subsection{3D Shape Generative Network}
The goal of this module is to fuse query text and multi-modal prior knowledge for more accurate structure information representation, which includes four key parts: 1) Feature selection: We introduce the causal inference model to remove the unrelated structure information from the prior shape feature. 2) Prior fusion module: It learns the correlation between select prior knowledge and input textual information, combines them into the fused feature, and feeds into the generative network. 3) Generative network: By finetuning the pre-trained autoencoder with the guidance of prior knowledge, it projects text features into 3D feature space to achieve text-3D generation. 4) Diversity generation: It improves the diversity of the generation results within the proposed prior guided IMLE structure. We will detail these modules in the following subsections.

%

\subsubsection{Feature Selection}
Based on the query text $T$, we can obtain the related 3D shapes $P_s=\{s_1,s_2,...s_m\}$ and semantic attributes $P_a=\{a_1,a_2,...a_n\}$ as prior knowledge from the 3D shape knowledge graph $K$. However, we note that the related 3D shape $s_i$ either resembles the query text or matches only part of the information in the query text.
 
We hope to remove this unrelated information, save the useful information for the next fusion operation and guarantee the completeness of the fusion feature. Based on this analysis, we are inspired by \cite{yue2020interventional} and introduce the causal model into the fusion model.

\begin{figure}[t]
	\centering
	\includegraphics[width=1.0\linewidth]{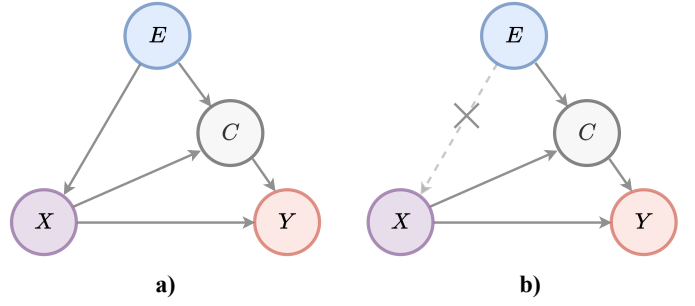}
  \caption{The causal graph, $X$ denotes the features of retrieval shapes extracted by the shape encoder $E_v$, $Y$ is the fusion feature of target shape, $E$ is the shape encoder, $C$ is the redundant information or unrelated feature in $X$ that act as the confounders in this causal model. }
  \label{backdoor}
\end{figure}

We first construct the causal graph as in Fig. \ref{backdoor}(a), where the nodes denote the data variables and the directed edges denote the (functional) causality. Here, $X=\{f_1,...,f_m\}$ denotes the features of the retrieval shapes extracted by the shape encoder $E_v$. $Y$ is the fusion feature of the target shape, which is constructed by $X$. $E=E_v$ is the shape encoder learned by the pre-trained model detailed in Section.3.1. $C$ is the redundant information or unrelated feature in $X$. $E\rightarrow X$ denotes the feature $X$ extracted by encoder $E$. $E\rightarrow C$ denotes the interference information is also extracted by $E$. $X\rightarrow C$ denotes that $C$ exists in $A$. $X\rightarrow Y \leftarrow C$ means that $Y$ can be directed by $X$ and also be influenced by $C$. In other words, the second way, $X\rightarrow C\rightarrow Y$, is inevitable because the encoder $E$ is not for feature fusion in the training step. Our goal is to find the true causality between $X$ and $Y$, and eliminate the interference of $C$. To pursue the true causality between $X$ and $Y$, we need to use the causal intervention $P(Y|do(X))$ instead of the likelihood $P(Y|X)$. 

In this paper, we propose using the backdoor adjustment to achieve $P(Y|do(X))$. The backdoor adjustment for the graph is as follows:
\begin{equation}
    P(Y|do(X))=\sum_d P(Y|X=x,C=g(X=x,E))P(E),
    \label{causal}
\end{equation}
where $g$ means that feature $X$ causes $C$ or $C$ is extracted from the prior feature $X$. $X$ comes from $\{f_1,...,f_m\}$ extracted from the related prior shapes, which includes the related structure information corresponding to the query text $T$. Meanwhile, it also includes unrelated information $C$. We apply the random sampling estimation to eliminate the influence of $C$.

First, we connect $\{f_1,...,f_n\}$ to construct $X=\{f_1:f_2:...:f_n\}\in R^{1\times nd}$. Suppose that $F$ is the index set of the feature dimensions of $X$. We divide $F$ into $n$ equal-size disjoint subsets. $F_i$ is the index of $X$. $g(x,E):=\{k|k\in F_i\cap I_t\}$, where $I_t$ is an index set whose corresponding absolute values in $X$ are larger than the threshold $t$. 
We set $t=e^{-3}$ in this paper. 
 
Here, we hope the final selected feature can contain as much information as possible about the structure information described by text $T$. Based on the pre-trained model $E_v$ and $E_t$, the text feature and shape feature belong to the same feature space. We think the selected feature should be similar to the target shape feature $f_s$ (ground truth). Thus, we define $Y\approx f_s$. Eq.\ref{causal} can be rewritten as:
\begin{equation}
    P(Y|do(X))=\frac{1}{n}\sum^n_{i=1} P(f_s|[X]_c),
    \label{causal2}
\end{equation}
where $c=\{k|k\in F_i\cap I_t\}$ is implemented as the index set defined above. $[X]_c$ is a feature selector that selects the dimensions of $x$ according to the index set $c$. $n$ is the number of samplings. $i$ is the i-th sampling. Here, we add one MLP layer to handle the selected feature $[X]_c$. The process can be defined as $x'_i=J_i([X]_c,w_i)$. $w_i$ is the parameter of the MLP layer. Based on this design, the final loss function can be written as:
\begin{equation}
    L=\frac{1}{n}\sum^n_{i=1} log(\frac{exp(f_s\cdot {x'}_i^T)}{\sum_{j=1}^n exp(f_s\cdot {x'}_j^T)}).
\end{equation}
By optimization, we will obtain $n$ number of optimization function $J$. For the shape prior knowledge $X=\{f_1:f_2:...:f_n\}$, we can obtain the processed and the selected features $F'_p=\{x'_1,...,x'_n\}$ as the input of the prior fusion module. 

\subsubsection{Prior Fusion Module}
\begin{figure}[t]
	\centering
	\includegraphics[width=1\linewidth]{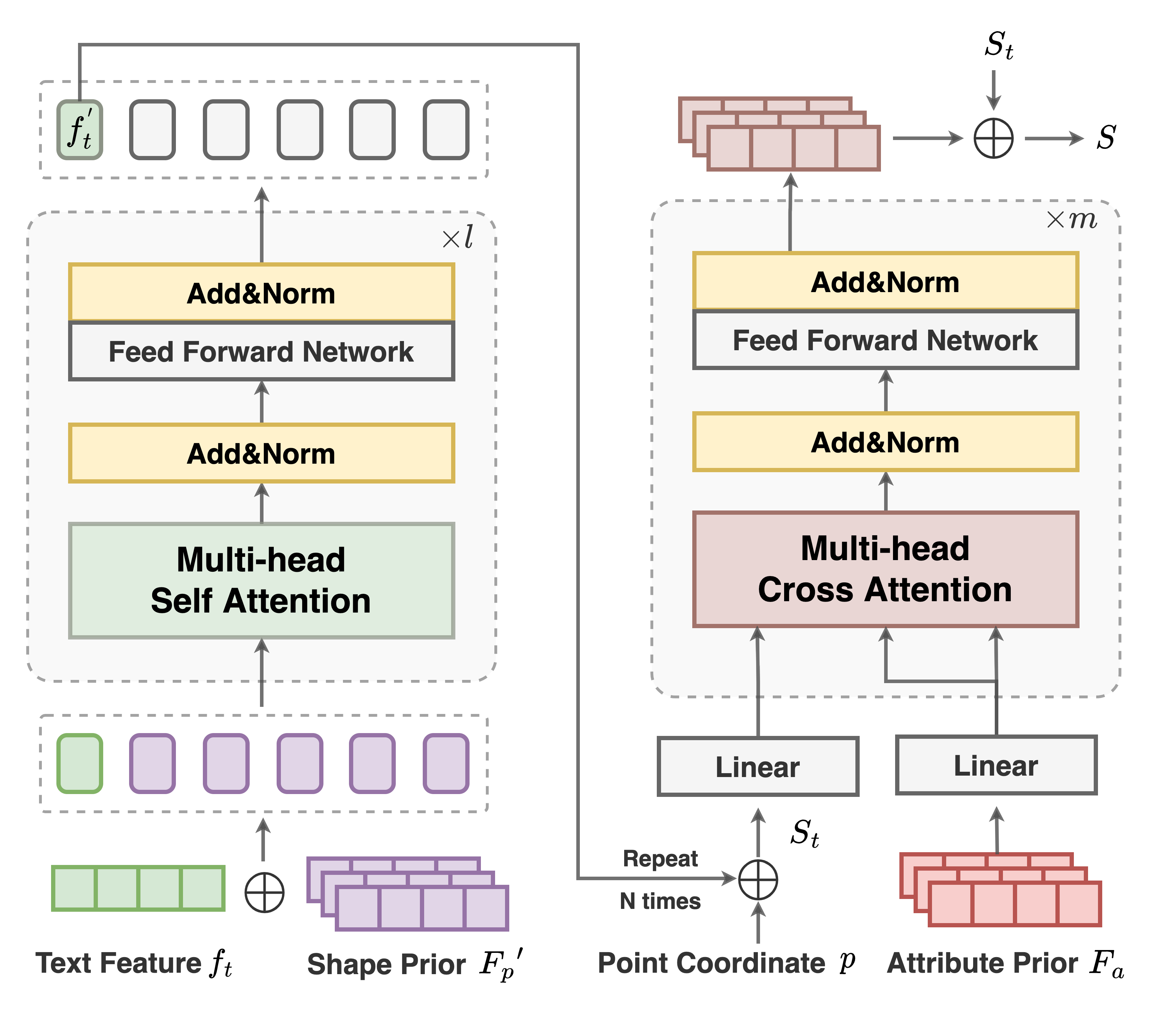}
  \caption{The network structure of the Prior Fusion Module(PFM). The left part fuses the shape prior information, which enriches the text feature with 3D information. The right part is used to fuse the prior attribute information.}
  \label{fusion}
\end{figure}

The PFM hierarchically integrates $F'_p=\{x'_1,...,x'_n\}$ and $F_a$ with $f_t$ in two steps. For each step, the calculation process is based on the stacked transformer blocks. Specifically, each layer of the transformer block has a multi-head attention module and a position-wise feed-forward network (FFN). The first step is to update $f_t$ with shape priors $F'_p$, setting $F_t^1 = \{f_t \oplus F'_p\}$ as the initial input sequence of the text feature and the selected shape prior. $F_t^i$ is the input feature of the $i_{th}$ layer, and the calculation process of each layer can be written as:
\begin{equation}
  \begin{aligned}
  Q &= W^Q \cdot F_t^{i-1}, K = W^K \cdot F_t^{i-1}, V = W^V \cdot F_t^{i-1} \\
  F_t^i &=Multihead(Q,K,V) \\
  F_t^i &=F F N(F_t^i),
  \end{aligned}
\end{equation}
where $i$ is the index of the transformer layers. Finally, in the last $l_{th}$ layer, we can obtain the updated text feature as $f_t^{\prime}$. This step aims to leverage the attention mechanism to learn the correlation between the text information and the shape priors, thus enriching the text feature with 3D information. Then, we adopt a similar idea to \cite{text2shapecvpr} to fuse attribute information in spatial feature space. Concatenating $f_t^{\prime}$ with the points position $p$ into the spatial feature $S_t = \{f_t^{\prime}\oplus p\} \in \mathbb{R}^{N\times(d+3)}$. Using fully-connected layers to convert $S_t$ and $F_a$ into $\hat{S_t}$, $\hat{F_a}$ with a lower favorable input dimension, similar to the first step, the attribute fusion step can be formulated as:
\begin{equation}
  \begin{aligned}
  Q &= W^Q \cdot \hat{S}_t^{j-1},K= W^K \cdot \hat{F_a}, V = W^V \cdot \hat{F_a} \\
  \hat{S_t^j} &=Multihead(Q,K,V) \\
  \hat{S_t^j} &=F F N(\hat{S_t^j}).
  \end{aligned}
\end{equation}

In the $m_{th}$ layer of the final part, the calculated $\hat{S_t^{\prime}} = \hat{S_t^{m}}$ will serve as extra information, concatenated with $S_t$ into $S=\{S_t\oplus\hat{S_t^{\prime}}\}$, which is the final fused feature used to feed into the 3D shape decoder. To adapt the dimension of the fused feature, 
the existing $D = \{D_s,D_c\}$ is extended to the dimension of the $\hat{S_t^{\prime}}$, and the extended 3D shape decoder is denoted as $D^{\prime} = \{D_s^{\prime},D_c^{\prime}\}$.

\subsubsection{Generative Network}
The basic framework of the generation network is shown in Fig. \ref{generation}(b,c), which includes the encoder $E_v$ and $E_t$ utilized for extracting the text and 3D shape features, respectively. The fusion module (PFM) fuses the query text information with prior knowledge. The decoder $D'$ is used to predict the final 3D shape model. 


To optimize the parameters of $E_v$, $PFM$, and $D\prime$ as well as to initialize the parameters of the network with the pre-trained checkpoint, we use the same $L_{ae}$ introduced above to renew training the autoencoder with prior knowledge guidance, which is formulated as:
\begin{equation}
  \begin{aligned}
    &L_{ae} = ||D_{s}^{'}(S) - I_s||_2 + ||D_{c}^{'}(S)\times I_s - I_c||_2.
  \end{aligned}
\end{equation}

For the framework to gain the ability to generate from text to 3D, an L2 norm-based regression loss $L_{reg}$ is applied to project text feature $f_t$ into 3D latent space.
\begin{equation}
  \begin{aligned}
    L_{reg} = ||f_t-f_s||_2 ,
  \end{aligned}
\end{equation}
where $f_t$ and $f_s$ are the extracted features of the text description $T$ and its corresponding 3D shape ground truth $V$. In the text-3D generation process, the $f_t$ can be directly used to synthesize the 3D model generation under the guidance of prior knowledge.
Finally, the optimization target of the entire generation network is:
\begin{equation}
  \begin{aligned}
    &L = {\lambda}L_{ae} + {(1-\lambda)}L_{reg},
  \end{aligned}
\end{equation}
where $\lambda$ is the weight parameter that controls the balance of the loss functions. We applied the Adam method\cite{adam} to optimize the generative network and obtain the parameters of $E_t$, $E_v$, $PFM$, and $D'$ for the text-3D generation.

\subsubsection{Diversity Generation}
Different people have different ideas. Therefore, the same text description should produce diverse shapes. 
To achieve the diverse shape generation results from the same text description, we adopt a similar idea with\cite{text2shapecvpr} by applying an IMLE\cite{imle} (implicit maximum likelihood estimation)-based latent generator $G$ to the extracted text features $f_t$ for randomness.
Here, given a set of random noise $Z = \{z_1,z_2\dots z_l\}$, the perturbed feature is formulated as $F'=G(f_t,Z)=\{f'_1,f'_2,\dots,f'_l\}$.
However, the original IMLE process has the limitation that it is challenging to generate a sufficiently large number of samples.
The reason for this is that the optimized process minimizes the distance between $F'$ and the ground truth $f_s$, which would result in no more significant changes from the random noise. This conclusion is supported by the final example\cite{text2shapecvpr}. 

\begin{figure*}[t]
	\centering
	\includegraphics[width=1\linewidth]{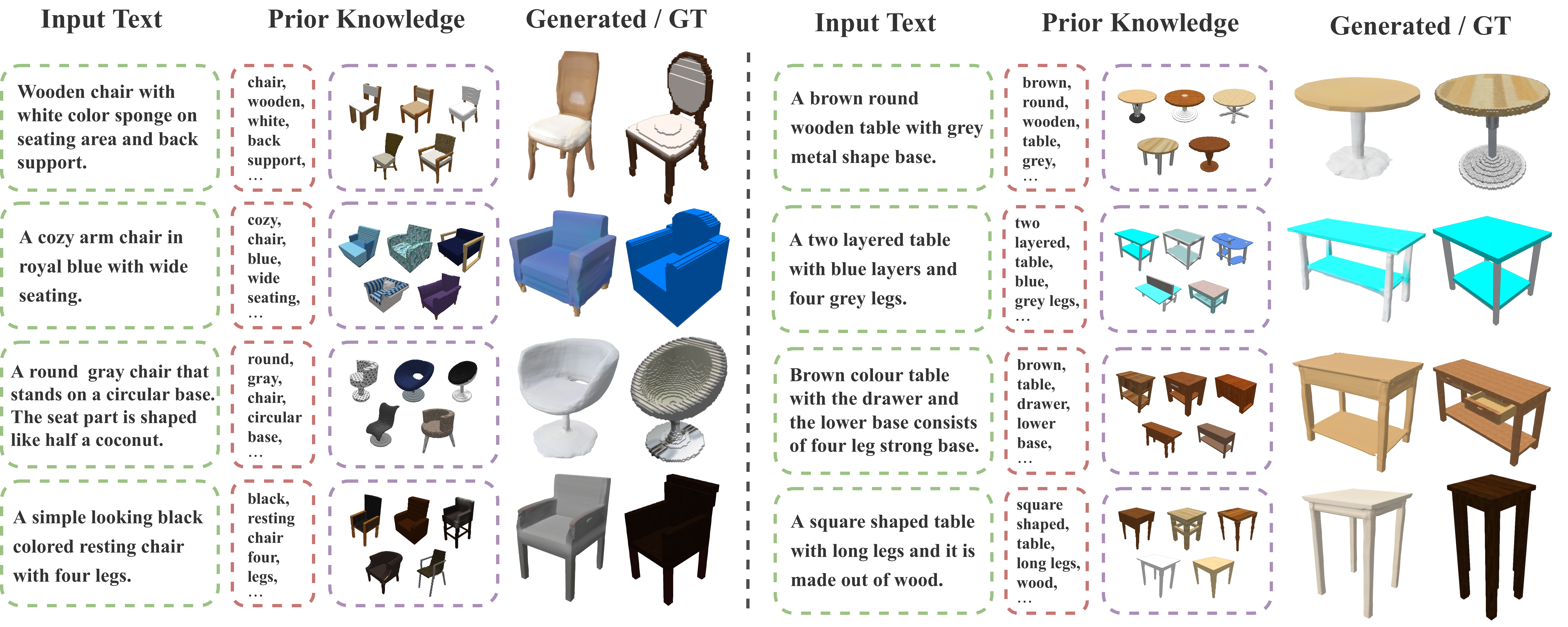}
  \caption{Several text-guided generation results. The models generated by our method basically contain the specific shape descriptions described in the text. The prior knowledge provided by the knowledge graph here provides certain supplementary information to ensure the similarity of the generative models.}
  \label{result-prior}
\end{figure*}

To overcome this difficulty, we introduce the shape of prior knowledge from the knowledge graph $G$ to increase the diversity. We achieve the related shape priors $F_p=\{f_p^1,f_p^2,...,f_p^m\}$ based on the query text $T$. Then, we resample a number of reference features $F_g=\{f_g^1,f_g^2,\dots,f_g^h\}$ using a linear interpolation function, which is calculated as:
\begin{equation}
    F_g = f_t+\frac{(F_p-f_t)}{\sigma}\cdot\eta,
\end{equation}
where $\sigma$ and $\eta$ control the range and step of the interpolation function, and $f_t$ is the feature of $T$. The sampled $F_g$ is an additional optimization objective, not $f_s$. For each perturbed feature $f'$, we reselect its optimization target by calculating its cosine similarity between $F_g$. The process is marked as follows:
\begin{equation}
    f_{target}=\arg\min_{i=1,...,h}d(G_{\phi}(f_t,Z),f_g^i),
\end{equation}
where $d$ is the distance metric, $\phi$ is the weights of the generator $G$. The goal is to find the optimization target $f_{target}$ from $F_g$.
The final optimized loss function can be written as:
\begin{equation}
    L_G=\min_{k=1,...,l}||G_{\phi}(f_t,z_k)-f_{target}||^2_2.
\end{equation}

In the optimization process, we first need to fix the parameter $\phi$ to find $f_{target}$. Then, we optimize $G$ based on the new target information. Here, $F_g$ provides a richer reference in the training process. The related experiments also demonstrate that our method can produce more variable models.

\section{Experiments}
To evaluate the effectiveness of the proposed framework, we carried out a series of experiments. At the beginning of this section, we introduce the dataset details and the experiment settings. In Sec. 4.3, we visualize several generated results and make comparisons with the SOTA methods. To further verify the effectiveness of each proposed module, we conducted ablation studies and comparative experiments, as shown in Sec. 4.4. 

\subsection{Dataset}
We conduct the experiments on the text-3D dataset in \cite{text2shape}, which consists of a primitive subset and a ShapeNet\cite{shapenet} subset. We use the ShapeNet subset to build the text-3D knowledge graph and conduct experiments. It contains 6,521 chairs and 8,378 tables of 3D volumes. Five text captions are presented for each 3D shape. To conduct the experiments, we follow the same training/validation/testing split as in the previous related works\cite{text2shape}.

\subsection{Experimental Settings}
We implement our proposed framework on PyTorch and use an Nvidia Tesla A40 GPU to complete all experiments.
To pre-train the representation module, we first train the autoencoder in the output resolution of $16^3$, then further refine the parameters in $16^3$, and finally the $L_{joint}$ is utilized to optimize the text encoder. The process is optimized with an Adam optimizer with a beta of 0.99, an initial learning rate of $10^{-4}$.

Based on the data pre-processing, we construct the knowledge graph and build the training data. For the text-3D generation network, we train the network end-to-end by initializing the network with the pre-trained parameters. To make the training process stable, we adjust the weight of each proposed loss function to $\alpha=1, \beta=0.1$. Similarly, we use Adam optimizer with a beta of 0.99, initial learning of $10^{-5}$ to train the network. With a batch size of 32, it takes up about 42 GB of GPU memory and takes around 50 hours to train 400 epochs. We select the trained models with the lowest validation loss for visualization and calculate the metrics for quantitative analysis.

\subsection{Comparison with the SOTA Methods}

In this section, two existing approaches, Text2Shape\cite{text2shape} and Text-Guided 3D\cite{text2shapecvpr} are served as the compared methods. Followed by these methods, we adopt the same evaluation metrics to make quantitative comparisons.
Evaluation metrics include 1) IOU (Intersection Over Union): which is used to measure the shape similarity between two 3D shapes; 2) EMD (Earth Mover's Distance): which is used to measure the color similarity; 3) IS (Inception Score\cite{inception}): it is used to measure the diversity and quality of the generated shapes; 4) Acc(Classification accuracy): it measures the accuracy of generated 3D shapes in the correct category, which is calculated by a pre-trained 3D shape classifier.
The final experimental results are shown in Table.\ref{compare-table}. From these results, our approach achieves the best performance. In IOU, EMD, and ACC, our approach obtains 2.3\%, 0.12\%, and 1.2\% improvements, respectively. We think there are some reasons as follows:

\begin{table}[t]
  \centering
  \renewcommand\arraystretch{1.8}
  \caption{Quantitative comparison with the SOTA methods: following the prior works, we identically report IOU, EMD, IS and Accuracy (Acc. (\%)) metrics to serve as the comparable quantitative evaluations.}
  \begin{tabular}{l|cccc}
  \hline
    Method      & IOU$\uparrow$           & EMD$\downarrow$            & IS$\uparrow$            & Acc$\uparrow$            \\ \hline
    Text2Shape\cite{text2shape}  & 9.64                     & 0.4443                   & 1.96                    & 97.37          \\
    Text-Guided\cite{text2shapecvpr} & 12.21                    & 0.2071                   & 1.97                    & 97.48          \\ \hline
    Ours        & \pmb{14.22}            & \pmb{0.1742}           & \pmb{1.97}           & \pmb{98.15} \\ \hline
    \end{tabular}
    \label{compare-table}
\end{table}

\begin{figure}[t]
	\centering
	\includegraphics[width=1\linewidth]{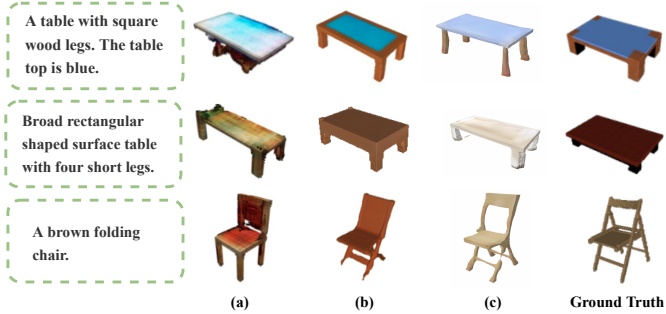}
  \caption{Generation results by Text2shape\cite{text2shape}(a) $vs$. Text-Guided\cite{text2shapecvpr}(b) $vs$. ours(c) $vs$. GT. By comparison, our algorithm obtains the 3D models that better fit the text description.}
  \label{compare}
\end{figure}

\begin{itemize}
\item Text2Shape\cite{text2shape} first applies a triplet loss to guide the cross-modal feature learning in a joint space, and then directly predicts the 3D volume with the text feature as conditional input. Due to the inadequate 3D information and the unstable training process of GAN, the synthesis generation is low-quality. 
\item Text-Guided\cite{text2shapecvpr} achieves the better improvement compared to Text2Shape. However, it only pays attention to the information alignment between the text-3D pair of the training data and ignores the training difficulties caused by the flexibility and ambiguity of the natural language. It tends to generate 3D shapes that are similar to the ground truth 3D data.
\item Our approach achieves the best performance. In our method, the introduction of prior knowledge can supplement additional information for text description to help generate 3D shapes. In addition, the introduction of the causal inference model eliminates the irrelevant information in the related shapes, so as to provide prior knowledge with higher confidence, which can greatly enhance the final generation performance.
\end{itemize}

Fig.\ref{result-prior} shows some generation results conditioned with the input text description and the retrieved prior knowledge. Fig.\ref{compare} compares the generation qualities of our framework with some classic generation methods. From the visualization results, we make the following observations: 
\begin{itemize}
    \item As seen in Figure 7, most of the 3D models retrieved by the proposed method can semantically match the input text descriptions, and they can supply the generative network with supplementary 3D information for more accurate shape generation. For example, in the last example, the generative network may be difficult to understand how the textual information ``with the drawer'' can be represented for the 3D shape. The retrieved 3D shapes can help the generative network to determine the basic structural characteristics of the generation target, to ensure the final generation quality.
    \item From Fig.\ref{compare}, our approach provides a more accurate model than the other two methods. The introduction of additional prior knowledge ensures that our method can generate 3D shapes that better match the input description. Especially in the last example, when faced with inadequate textual information, with only a few attributes ``brown'' and  ``folding chair'', the two comparison methods only generate approximate appearance, while our methods can produce a more accurate 3D shape.
\end{itemize}

  \begin{figure}[t] 
    \centering
    \includegraphics[width=1\linewidth]{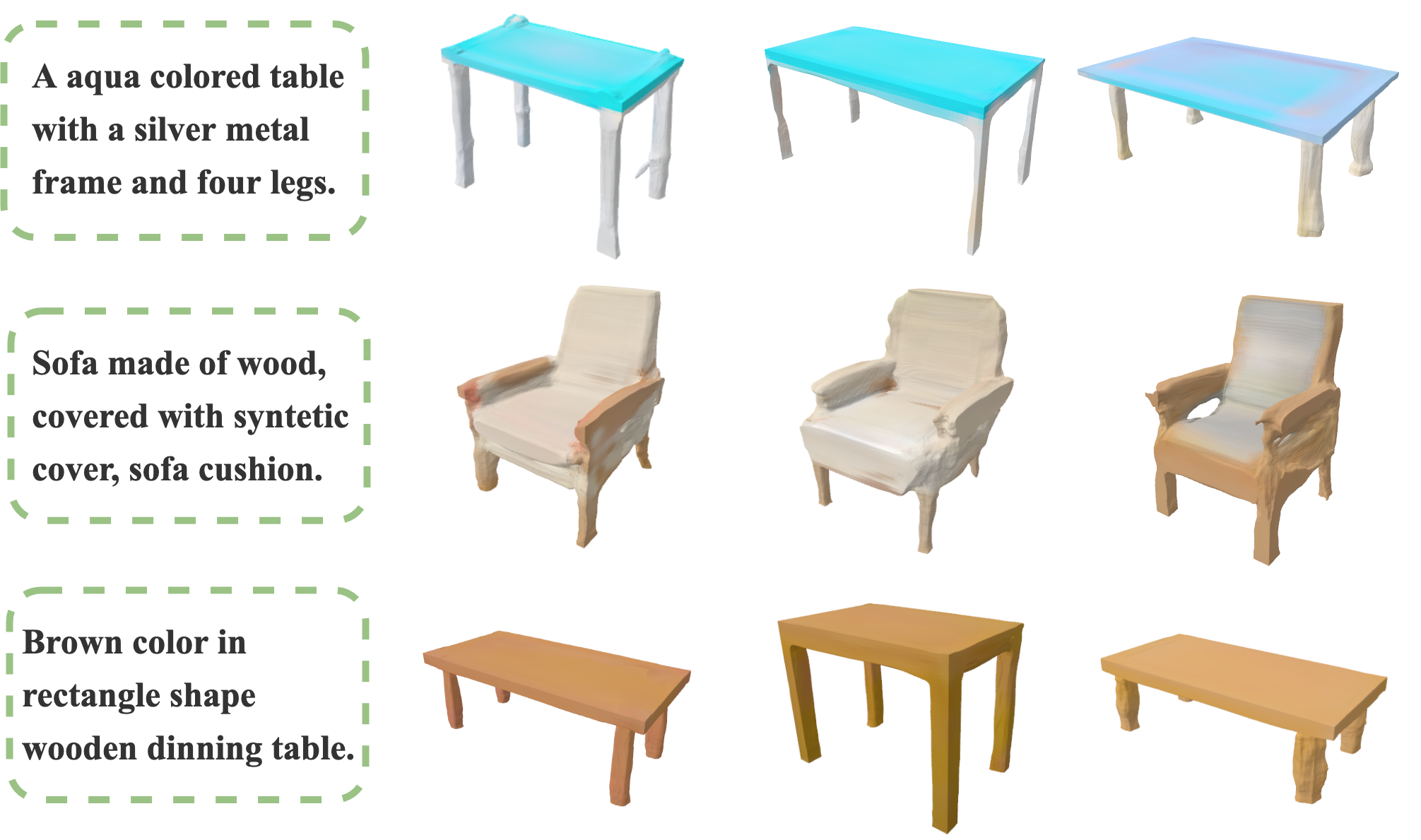}
    \caption{Visualization of several diversifying generation results. These generative models have more variation regarding the ground truth shape, which can meet more users' requirements.} 
    \label{diverse}
  \end{figure}
  \subsection{Diversity Generation}
  To evaluate the performance of the diversity generator $G$. Some diversity generation results are shown in Fig. \ref{diverse}. By applying an IMLE-based latent generator, the text feature $f_t$ can be converted into a different perturbed $f'$. Using the trained $D'$, diverse 3D shapes corresponding to the single input text can be generated. 
  
  From Fig.\ref{diverse}, we find more variations in the results obtained by our approach compared to the Text-guide 3D method. For example, our approach can achieve tables of different heights in the third example. Meanwhile, these generative models have more variation in the ground truth shape. In practice, it can meet the requirements of more users. These results also demonstrate the performance of our diversity generator $G$.

 \subsection{Ablation Studies} 
We conducted extensive ablation studies to verify the effectiveness of each proposed module. The experimental results are shown in Table. \ref{ablation1} and Table. \ref{ablation2}. In this section, we introduce each experiment setting in our ablation studies and analyze the effectiveness of each proposed module.

For a quantitative evaluation, we adopt the same metrics as the ablation study settings in \cite{text2shapecvpr}, which applied Point Score (PS) and Frechet Point Distance(FPD) to evaluate the qualities of generated 3D shapes. In our cases, we sample the 3D shapes in ShapeNet\cite{shapenet} into colored point clouds of 55 classes, then train the classification-based inception network for PS and FPD calculations. The R-Precision\cite{AttnGAN} is also applied here to measure the correlation between the input text description and generated 3D shapes. We use the text and shape encoder trained by $L_{joint}$ in the proposed representation module to extract features of the original text description in the test set and the generated 3D shapes. For each generated shape, we use the extracted feature to retrieve the related text, and calculate the retrieval accuracy in the top 20 results as its R-Precision.
We will detail our observations in the next subsections. 

\begin{table}[t]
  \centering
  \renewcommand\arraystretch{1.6}
  \caption{Quantitative experiment results of the ablation studies: we report IOU, PS, FPD, and R-Precision (R-P.(\%) to quantitatively evaluate the effectiveness of the applied loss functions and prior knowledge.}
  \begin{tabular}{c|p{0.9cm}<{\centering}p{0.9cm}<{\centering}p{0.9cm}<{\centering}p{0.9cm}<{\centering}}
  \hline
    Method      & IOU$\uparrow$           & PS$\uparrow$            & FPD$\downarrow$            & R-P$\uparrow$            \\ \hline
    baseline  & 9.23  & 2.54 & 234.91 & 9.34         \\
    $+L_{reg}$ & 12.20 & 2.93 & 111.43 & 38.00          \\
    $+L_{reg}\ +L_{ae}$  & 13.13 & 3.14 & 51.12 & 41.84 \\ \hline
    +attribute prior  & 13.34 & 3.18 & 46.54 & 43.35         \\
    +shape prior    & 14.07 & 3.22 & 40.35 & 42.32          \\
    +shape prior(Causal) & 14.13 & 3.27 & 36.22 &  43.85        \\
    +both prior(Causal)  & \pmb{14.22} & \pmb{3.35} & \pmb{30.71} & \pmb{45.70} \\ \hline
    \end{tabular}
    \label{ablation1}
\end{table}

\begin{table}[t]
  \centering
  \renewcommand\arraystretch{1.6}
  \caption{Quantitative comparison of the prior fusion methods: the metrics IOU, PS, FPD, and R-Precision (R-P.(\%) are also used here to evaluate the performance of different prior fusion methods.}
  \begin{tabular}{c|p{0.9cm}<{\centering}p{0.9cm}<{\centering}p{0.9cm}<{\centering}p{0.9cm}<{\centering}}
  \hline
    Method      & IOU$\uparrow$           & PS$\uparrow$            & FPD$\downarrow$            & R-P$\uparrow$            \\ \hline
    Concatenate    & 12.74 & 3.08 & 45.73 & 41.31          \\
    Average Fusion & 13.32 & 2.96 & 74.28 & 40.77       \\
    +Ours(PFM)  & \pmb{14.07} & \pmb{3.22} & \pmb{40.35} & \pmb{42.32} \\ \hline
    \end{tabular}
    \label{ablation2}
\end{table}

\begin{figure}[t] 
	\centering
	\includegraphics[width=1\linewidth]{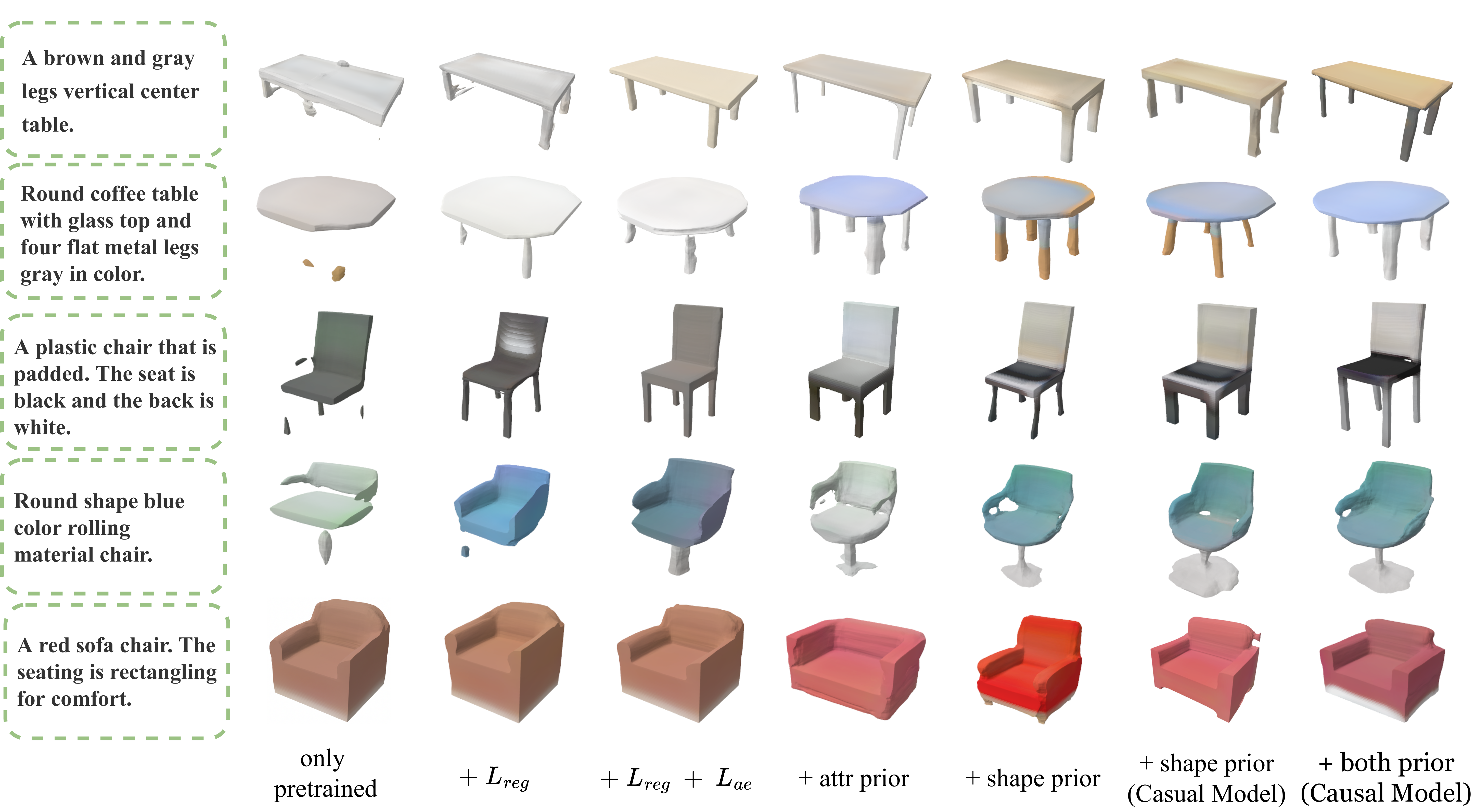}
  \caption{Visualization of ablation studies, which shows the effects of the introduction of each module. From the shown results we can find that: without the introduction of prior knowledge,$ +L_{reg}\ +L_{ae}$ can only generate roughly matching shapes, and the utilization of the attribute and shape priors can enrich the details from different perspectives. The experiment setting "+attr prior" makes the generated shapes more semantically compatible with the input text, and "+shape prior" introduces more accurate geometrics and richer colors to the generated shapes. Finally, the introduction of the "+causal model" provides a better generation based on the introduction of both prior knowledge.} 
  \label{ablation-results}
\end{figure}

\subsubsection{Loss Function}
First, we conducted experiments to verify the influence of each applied loss function to determine the circumstances necessary for the framework.
\begin{itemize}
  \item ``Baseline'' means that we directly use the text encoder $E_v$ and shape decoder $D'$ pre-trained in the step of learning joint representation. It achieves the worst results. 
  \item ``$+L_{reg}$'' means that we only optimize the text encoder $E_t$ to project the text feature into the learned autoencoder space. The aim of $L_{reg}$ is to constrain the encoded text feature $f_t$ to be similar to the extracted feature $f_s$ of their corresponding 3D shape.
  \item ``$+L_{reg}+L_{ae}$'' indicates that we further applied $L_{ae}$ to train the entire framework end-to-end. In this experimental setting, the trained autoencoder is further finetuned with the joint of textual information. It achieves better performance compared with ``$+L_{reg}$''.

\end{itemize}

\subsubsection{Prior Knowledge}
Given the input text description, we retrieve the prior knowledge of the related attributes and the 3D shapes to assist with the text-3D generation in the proposed framework.
In the ablation experiment of this part, the previous setting ``$+L_{reg}+L_{ae}$'' can be seen as the baseline that does not use any prior knowledge in the entire process. The experimental settings include three parts:
\begin{itemize}
  \item ``$+attribute~prior$'' means that we only add the attribute prior $F_a$ into the training process, and use the proposed prior fusion module to update the spatial features. By constructing the attention map between the spatial features and the attribute information, the generated shapes achieve better qualities both in their generated structures and colors, which is also reflected in the increase of quantitative metrics.
  \item ``$+shape~prior$'' means that we only add the retrieved shape prior $F_p$ to update the extracted text feature $f_t$. The aim of introducing the shape prior is to supplement the lack of specific geometric information of $f_t$ in the high-level feature space. The number of utilized shapes is limited by $5$, which is the default setting of our proposed framework. From the shown results, we can see that the introduction of the shape prior can also improve the generation quality.
   \item ``$+shape~prior$ (Causal Model)'' means that we add the causal model to select useful features for the next feature fusion operation. The number of utilized shapes is also limited by $5$. From the results, the causal model brings a significant improvement, which demonstrates that the causal model can effectively reduce the unrelated shape information from the shape prior knowledge and improve the performance of the fusion feature. 
  \item ``$+both~prior$ (Causal Model)'' means that we applied both the attributes and the shape prior knowledge into the generation framework, which is also the final method we introduced in the main paper. The introduction of two kinds of prior knowledge can achieve mutual compatibility and make the best generation qualities.
\end{itemize}

\subsubsection{Prior Fusion Method}
The way to integrate the retrieved prior knowledge is also critical. To best leverage the correlations between the text and the retrieved 3D shapes, we designed the prior fusion transformer (PFM) to update the text feature with prior knowledge. To verify its effectiveness, similar to prior works\cite{3Drec_mem}, we set two fusion methods as comparison methods. The ``concatenate'' means we simply connect the shapes' feature with the text feature, and use a fully connected layer to transform the fusion feature into a favorable dimension. The ``Average Fusion'' means that we directly use the average pooling function to fuse the text features with prior knowledge. The experimental results are shown in Table. \ref{ablation2}, and the proposed prior fusion modules perform better.

From these experimental results, we find that the introduction of prior knowledge greatly improves the performance of the generation model. The shape of prior knowledge brings a larger improvement, which also demonstrates that the shape of the prior knowledge effectively makes up for the lack of structure information and improves the final performance. The causal model also plays a key role in the step of feature fusion, which provides a more plausible explanation for the increase in performance. The corresponding experimental results also demonstrate its superiority. The PFM modules applied to the transformer structure can reduce the effect of redundant information and improve the performance of the fused feature. 

\section{Conclusion}
In this paper, we proposed a novel text-3D generation model with the utilization of prior knowledge. Here, we first proposed a novel 3D shape knowledge graph to bridge the gap between text and 3D models. We save and achieve richer and more accurate prior knowledge like human beings. Then, we proposed a novel casual model to select useful and related features and remove the unrelated structure information from the searched shapes' prior knowledge. Combined with the information fusion model of this paper, we achieve an effective fusion feature as the input of the 3D generation model. The final experimental results demonstrated that our approach significantly improves 3D mode generation quality and performs favorably against the SOTA methods on the Text2shape\cite{text2shape} datasets.

From these experiments, we find that the 3D shape knowledge graph plays one key role in this work, which saves the correlation between text and 3D shapes. If we introduce more data and increase the size of the knowledge graph, it provides more accurate related prior knowledge like a wiser old man to help the target 3D generation. In future work, we will expand the existing database to increase the size of the knowledge graph. Meanwhile, the causal model plays a very important role in the selection of features. The related experiments also demonstrate this conclusion. In future work, we plan to introduce more partial structure information to structure causal graphs and optimization mechanisms. The generation model can more intelligently filter and utilize prior knowledge.

\section*{Acknowledgments}
This work was supported in part by the National Natural Science Foundation of China (62272337, 61872267) and the Natural Science Foundation of Tianjin (16JCZDJC31100, 16JCZDJC31100).
\normalem
\bibliographystyle{IEEEtran}
\bibliography{egbib}

\begin{thebibliography}{10}
\providecommand{\url}[1]{#1}
\csname url@samestyle\endcsname
\providecommand{\newblock}{\relax}
\providecommand{\bibinfo}[2]{#2}
\providecommand{\BIBentrySTDinterwordspacing}{\spaceskip=0pt\relax}
\providecommand{\BIBentryALTinterwordstretchfactor}{4}
\providecommand{\BIBentryALTinterwordspacing}{\spaceskip=\fontdimen2\font plus
\BIBentryALTinterwordstretchfactor\fontdimen3\font minus
  \fontdimen4\font\relax}
\providecommand{\BIBforeignlanguage}[2]{{%
\expandafter\ifx\csname l@#1\endcsname\relax
\typeout{** WARNING: IEEEtran.bst: No hyphenation pattern has been}%
\typeout{** loaded for the language `#1'. Using the pattern for}%
\typeout{** the default language instead.}%
\else
\language=\csname l@#1\endcsname
\fi
#2}}
\providecommand{\BIBdecl}{\relax}
\BIBdecl

\bibitem{3Dgen_3Dvaegan}
J.~Liu, F.~Yu, and T.~Funkhouser, ``Interactive 3d modeling with a generative
  adversarial network,'' in \emph{2017 International Conference on 3D Vision
  (3DV)}.\hskip 1em plus 0.5em minus 0.4em\relax IEEE, 2017, pp. 126--134.

\bibitem{3Dgen_grass}
J.~Li, K.~Xu, S.~Chaudhuri, E.~Yumer, H.~Zhang, and L.~Guibas, ``Grass:
  Generative recursive autoencoders for shape structures,'' \emph{ACM
  Transactions on Graphics (TOG)}, vol.~36, no.~4, pp. 1--14, 2017.

\bibitem{3Dgen_spgan}
R.~Li, X.~Li, K.-H. Hui, and C.-W. Fu, ``Sp-gan: Sphere-guided 3d shape
  generation and manipulation,'' \emph{ACM Transactions on Graphics (TOG)},
  vol.~40, no.~4, pp. 1--12, 2021.

\bibitem{3dgen-substructure}
C.~Zhu, K.~Xu, S.~Chaudhuri, R.~Yi, and H.~Zhang, ``Scores: Shape composition
  with recursive substructure priors,'' \emph{ACM Transactions on Graphics
  (TOG)}, vol.~37, no.~6, pp. 1--14, 2018.

\bibitem{3Drec_3DR2N2}
C.~B. Choy, D.~Xu, J.~Gwak, K.~Chen, and S.~Savarese, ``3d-r2n2: A unified
  approach for single and multi-view 3d object reconstruction,'' in
  \emph{European conference on computer vision}.\hskip 1em plus 0.5em minus
  0.4em\relax Springer, 2016, pp. 628--644.

\bibitem{3Drec_pix2vox}
H.~Xie, H.~Yao, X.~Sun, S.~Zhou, and S.~Zhang, ``Pix2vox: Context-aware 3d
  reconstruction from single and multi-view images,'' in \emph{Proceedings of
  the IEEE/CVF international conference on computer vision}, 2019, pp.
  2690--2698.

\bibitem{3Drec_pix2vox++}
H.~Xie, H.~Yao, S.~Zhang, S.~Zhou, and W.~Sun, ``Pix2vox++: Multi-scale
  context-aware 3d object reconstruction from single and multiple images,''
  \emph{International Journal of Computer Vision}, vol. 128, no.~12, pp.
  2919--2935, 2020.

\bibitem{3Drec_pami}
X.~Zhang, R.~Ma, C.~Zou, M.~Zhang, X.~Zhao, and Y.~Gao, ``View-aware
  geometry-structure joint learning for single-view 3d shape reconstruction,''
  \emph{IEEE Transactions on Pattern Analysis and Machine Intelligence}, 2021.

\bibitem{pix2mesh++}
N.~Wang, Y.~Zhang, Z.~Li, Y.~Fu, H.~Yu, W.~Liu, X.~Xue, and Y.-G. Jiang,
  ``Pixel2mesh: 3d mesh model generation via image guided deformation,''
  \emph{IEEE transactions on pattern analysis and machine intelligence},
  vol.~43, no.~10, pp. 3600--3613, 2020.

\bibitem{3Dscene_pami}
X.~Zhang, R.~Ma, C.~Zou, M.~Zhang, X.~Zhao, and Y.~Gao, ``View-aware
  geometry-structure joint learning for single-view 3d shape reconstruction,''
  \emph{IEEE Transactions on Pattern Analysis and Machine Intelligence}, 2021.

\bibitem{3Dscene_m2}
S.~Liu, Y.~Hu, Y.~Zeng, Q.~Tang, B.~Jin, Y.~Han, and X.~Li, ``See and think:
  Disentangling semantic scene completion,'' \emph{Advances in Neural
  Information Processing Systems}, vol.~31, 2018.

\bibitem{3Dscene_m3}
S.~Li, C.~Zou, Y.~Li, X.~Zhao, and Y.~Gao, ``Attention-based multi-modal fusion
  network for semantic scene completion,'' in \emph{Proceedings of the AAAI
  Conference on Artificial Intelligence}, vol.~34, no.~07, 2020, pp.
  11\,402--11\,409.

\bibitem{3Drec_sketch}
Z.~Lun, M.~Gadelha, E.~Kalogerakis, S.~Maji, and R.~Wang, ``3d shape
  reconstruction from sketches via multi-view convolutional networks,'' in
  \emph{2017 International Conference on 3D Vision (3DV)}.\hskip 1em plus 0.5em
  minus 0.4em\relax IEEE, 2017, pp. 67--77.

\bibitem{3Drec_sketch2}
L.~Wang, C.~Qian, J.~Wang, and Y.~Fang, ``Unsupervised learning of 3d model
  reconstruction from hand-drawn sketches,'' in \emph{Proceedings of the 26th
  ACM international conference on Multimedia}, 2018, pp. 1820--1828.

\bibitem{3Drec_sketch3}
S.-H. Zhang, Y.-C. Guo, and Q.-W. Gu, ``Sketch2model: View-aware 3d modeling
  from single free-hand sketches,'' in \emph{Proceedings of the IEEE/CVF
  Conference on Computer Vision and Pattern Recognition}, 2021, pp. 6012--6021.

\bibitem{nie2019hgan}
W.~Nie, W.~Wang, A.~Liu, J.~Nie, and Y.~Su, ``Hgan: Holistic generative
  adversarial networks for two-dimensional image-based three-dimensional object
  retrieval,'' \emph{ACM Transactions on Multimedia Computing, Communications,
  and Applications (TOMM)}, vol.~15, no.~4, pp. 1--24, 2019.

\bibitem{3drev1}
L.~Jing, E.~Vahdani, J.~Tan, and Y.~Tian, ``Cross-modal center loss for 3d
  cross-modal retrieval,'' in \emph{Proceedings of the IEEE/CVF Conference on
  Computer Vision and Pattern Recognition (CVPR)}, June 2021, pp. 3142--3151.

\bibitem{3drev2}
M.-X. Lin, J.~Yang, H.~Wang, Y.-K. Lai, R.~Jia, B.~Zhao, and L.~Gao, ``Single
  image 3d shape retrieval via cross-modal instance and category contrastive
  learning,'' in \emph{Proceedings of the IEEE/CVF International Conference on
  Computer Vision (ICCV)}, October 2021, pp. 11\,405--11\,415.

\bibitem{text2shape}
K.~Chen, C.~B. Choy, M.~Savva, A.~X. Chang, T.~Funkhouser, and S.~Savarese,
  ``Text2shape: Generating shapes from natural language by learning joint
  embeddings,'' in \emph{Asian conference on computer vision}.\hskip 1em plus
  0.5em minus 0.4em\relax Springer, 2018, pp. 100--116.

\bibitem{reed1}
S.~Reed, Z.~Akata, H.~Lee, and B.~Schiele, ``Learning deep representations of
  fine-grained visual descriptions,'' in \emph{Proceedings of the IEEE
  conference on computer vision and pattern recognition}, 2016, pp. 49--58.

\bibitem{reed2}
S.~E. Reed, Z.~Akata, S.~Mohan, S.~Tenka, B.~Schiele, and H.~Lee, ``Learning
  what and where to draw,'' \emph{Advances in neural information processing
  systems}, vol.~29, 2016.

\bibitem{text2shapecvpr}
Z.~Liu, Y.~Wang, X.~Qi, and C.-W. Fu, ``Towards implicit text-guided 3d shape
  generation,'' in \emph{Proceedings of the IEEE/CVF Conference on Computer
  Vision and Pattern Recognition}, 2022, pp. 17\,896--17\,906.

\bibitem{xiong2021knowledge}
H.~Xiong, S.~Wang, M.~Tang, L.~Wang, and X.~Lin, ``Knowledge graph question
  answering with semantic oriented fusion model,'' \emph{Knowledge-Based
  Systems}, vol. 221, p. 106954, 2021.

\bibitem{3Dgen_treegan}
D.~W. Shu, S.~W. Park, and J.~Kwon, ``3d point cloud generative adversarial
  network based on tree structured graph convolutions,'' in \emph{Proceedings
  of the IEEE/CVF international conference on computer vision}, 2019, pp.
  3859--3868.

\bibitem{3Dgen_diffusion}
S.~Luo and W.~Hu, ``Diffusion probabilistic models for 3d point cloud
  generation,'' in \emph{Proceedings of the IEEE/CVF Conference on Computer
  Vision and Pattern Recognition}, 2021, pp. 2837--2845.

\bibitem{3Dgen_diffusion2}
L.~Zhou, Y.~Du, and J.~Wu, ``3d shape generation and completion through
  point-voxel diffusion,'' in \emph{Proceedings of the IEEE/CVF International
  Conference on Computer Vision}, 2021, pp. 5826--5835.

\bibitem{pix2mesh}
N.~Wang, Y.~Zhang, Z.~Li, Y.~Fu, W.~Liu, and Y.-G. Jiang, ``Pixel2mesh:
  Generating 3d mesh models from single rgb images,'' in \emph{Proceedings of
  the European conference on computer vision (ECCV)}, 2018, pp. 52--67.

\bibitem{pix2mesh_pami}
J.~Tang, X.~Han, M.~Tan, X.~Tong, and K.~Jia, ``Skeletonnet: A
  topology-preserving solution for learning mesh reconstruction of object
  surfaces from rgb images,'' \emph{IEEE transactions on pattern analysis and
  machine intelligence}, 2021.

\bibitem{implicit-im-net}
Z.~Chen and H.~Zhang, ``Learning implicit fields for generative shape
  modeling,'' in \emph{Proceedings of the IEEE/CVF Conference on Computer
  Vision and Pattern Recognition}, 2019, pp. 5939--5948.

\bibitem{implicit-deepsdf}
J.~J. Park, P.~Florence, J.~Straub, R.~Newcombe, and S.~Lovegrove, ``Deepsdf:
  Learning continuous signed distance functions for shape representation,'' in
  \emph{Proceedings of the IEEE/CVF conference on computer vision and pattern
  recognition}, 2019, pp. 165--174.

\bibitem{implicit-disn}
Q.~Xu, W.~Wang, D.~Ceylan, R.~Mech, and U.~Neumann, ``Disn: Deep implicit
  surface network for high-quality single-view 3d reconstruction,''
  \emph{Advances in Neural Information Processing Systems}, vol.~32, 2019.

\bibitem{implicit-peoplesdf}
J.~Chibane, T.~Alldieck, and G.~Pons-Moll, ``Implicit functions in feature
  space for 3d shape reconstruction and completion,'' in \emph{Proceedings of
  the IEEE/CVF Conference on Computer Vision and Pattern Recognition}, 2020,
  pp. 6970--6981.

\bibitem{implicit-autosdf}
P.~Mittal, Y.-C. Cheng, M.~Singh, and S.~Tulsiani, ``Autosdf: Shape priors for
  3d completion, reconstruction and generation,'' in \emph{Proceedings of the
  IEEE/CVF Conference on Computer Vision and Pattern Recognition}, 2022, pp.
  306--315.

\bibitem{implicit-occnet}
L.~Mescheder, M.~Oechsle, M.~Niemeyer, S.~Nowozin, and A.~Geiger, ``Occupancy
  networks: Learning 3d reconstruction in function space,'' in
  \emph{Proceedings of the IEEE/CVF conference on computer vision and pattern
  recognition}, 2019, pp. 4460--4470.

\bibitem{implicit-template}
Z.~Zheng, T.~Yu, Q.~Dai, and Y.~Liu, ``Deep implicit templates for 3d shape
  representation,'' in \emph{{IEEE} Conference on Computer Vision and Pattern
  Recognition, {CVPR} 2021, virtual, June 19-25, 2021}.\hskip 1em plus 0.5em
  minus 0.4em\relax Computer Vision Foundation / {IEEE}, 2021, pp. 1429--1439.

\bibitem{implicit-deformation}
Y.~Deng, J.~Yang, and X.~Tong, ``Deformed implicit field: Modeling 3d shapes
  with learned dense correspondence,'' in \emph{Proceedings of the IEEE/CVF
  Conference on Computer Vision and Pattern Recognition}, 2021, pp.
  10\,286--10\,296.

\bibitem{flower}
M.-E. Nilsback and A.~Zisserman, ``Automated flower classification over a large
  number of classes,'' in \emph{2008 Sixth Indian Conference on Computer
  Vision, Graphics \& Image Processing}.\hskip 1em plus 0.5em minus 0.4em\relax
  IEEE, 2008, pp. 722--729.

\bibitem{COCO}
T.-Y. Lin, M.~Maire, S.~Belongie, J.~Hays, P.~Perona, D.~Ramanan,
  P.~Doll{\'a}r, and C.~L. Zitnick, ``Microsoft coco: Common objects in
  context,'' in \emph{European conference on computer vision}.\hskip 1em plus
  0.5em minus 0.4em\relax Springer, 2014, pp. 740--755.

\bibitem{yfcc100m}
B.~Thomee, D.~A. Shamma, G.~Friedland, B.~Elizalde, K.~Ni, D.~Poland, D.~Borth,
  and L.-J. Li, ``Yfcc100m: The new data in multimedia research,''
  \emph{Communications of the ACM}, vol.~59, no.~2, pp. 64--73, 2016.

\bibitem{gan-int-cls}
S.~Reed, Z.~Akata, X.~Yan, L.~Logeswaran, B.~Schiele, and H.~Lee, ``Generative
  adversarial text to image synthesis,'' in \emph{International conference on
  machine learning}.\hskip 1em plus 0.5em minus 0.4em\relax PMLR, 2016, pp.
  1060--1069.

\bibitem{stackgan}
H.~Zhang, T.~Xu, H.~Li, S.~Zhang, X.~Wang, X.~Huang, and D.~N. Metaxas,
  ``Stackgan: Text to photo-realistic image synthesis with stacked generative
  adversarial networks,'' in \emph{Proceedings of the IEEE international
  conference on computer vision}, 2017, pp. 5907--5915.

\bibitem{stackgan++}
H.~\vspace{0mm}Zhang, T.~Xu, H.~Li, S.~Zhang, X.~Wang, X.~Huang, and D.~N.
  Metaxas, ``Stackgan++: Realistic image synthesis with stacked generative
  adversarial networks,'' \emph{IEEE transactions on pattern analysis and
  machine intelligence}, vol.~41, no.~8, pp. 1947--1962, 2018.

\bibitem{GAN}
I.~Goodfellow, J.~Pouget-Abadie, M.~Mirza, B.~Xu, D.~Warde-Farley, S.~Ozair,
  A.~Courville, and Y.~Bengio, ``Generative adversarial networks,''
  \emph{Communications of the ACM}, vol.~63, no.~11, pp. 139--144, 2020.

\bibitem{AttnGAN}
T.~Xu, P.~Zhang, Q.~Huang, H.~Zhang, Z.~Gan, X.~Huang, and X.~He, ``Attngan:
  Fine-grained text to image generation with attentional generative adversarial
  networks,'' in \emph{Proceedings of the IEEE conference on computer vision
  and pattern recognition}, 2018, pp. 1316--1324.

\bibitem{SEGAN}
Z.~Li, T.~Zhang, P.~Wan, and D.~Zhang, ``Segan: structure-enhanced generative
  adversarial network for compressed sensing mri reconstruction,'' in
  \emph{Proceedings of the AAAI Conference on Artificial Intelligence},
  vol.~33, no.~01, 2019, pp. 1012--1019.

\bibitem{MirrorGAN}
T.~Qiao, J.~Zhang, D.~Xu, and D.~Tao, ``Mirrorgan: Learning text-to-image
  generation by redescription,'' in \emph{Proceedings of the IEEE/CVF
  Conference on Computer Vision and Pattern Recognition}, 2019, pp. 1505--1514.

\bibitem{DMGAN}
N.~Zheng, J.~Ding, and T.~Chai, ``Dmgan: Adversarial learning-based decision
  making for human-level plant-wide operation of process industries under
  uncertainties,'' \emph{IEEE Transactions on Neural Networks and Learning
  Systems}, vol.~32, no.~3, pp. 985--998, 2020.

\bibitem{RiFeGAN}
J.~Cheng, F.~Wu, Y.~Tian, L.~Wang, and D.~Tao, ``Rifegan: Rich feature
  generation for text-to-image synthesis from prior knowledge,'' in
  \emph{Proceedings of the IEEE/CVF conference on computer vision and pattern
  recognition}, 2020, pp. 10\,911--10\,920.

\bibitem{df-gan}
M.~Tao, H.~Tang, F.~Wu, X.-Y. Jing, B.-K. Bao, and C.~Xu, ``Df-gan: A simple
  and effective baseline for text-to-image synthesis,'' in \emph{Proceedings of
  the IEEE/CVF Conference on Computer Vision and Pattern Recognition}, 2022,
  pp. 16\,515--16\,525.

\bibitem{Dalle}
A.~Ramesh, M.~Pavlov, G.~Goh, S.~Gray, C.~Voss, A.~Radford, M.~Chen, and
  I.~Sutskever, ``Zero-shot text-to-image generation,'' in \emph{International
  Conference on Machine Learning}.\hskip 1em plus 0.5em minus 0.4em\relax PMLR,
  2021, pp. 8821--8831.

\bibitem{CogView}
M.~Ding, Z.~Yang, W.~Hong, W.~Zheng, C.~Zhou, D.~Yin, J.~Lin, X.~Zou, Z.~Shao,
  H.~Yang \emph{et~al.}, ``Cogview: Mastering text-to-image generation via
  transformers,'' \emph{Advances in Neural Information Processing Systems},
  vol.~34, pp. 19\,822--19\,835, 2021.

\bibitem{shapecaptioner}
Z.~Han, C.~Chen, Y.-S. Liu, and M.~Zwicker, ``Shapecaptioner: Generative
  caption network for 3d shapes by learning a mapping from parts detected in
  multiple views to sentences,'' in \emph{Proceedings of the 28th ACM
  International Conference on Multimedia}, 2020, pp. 1018--1027.

\bibitem{y2seq2seq}
Z.~Han, M.~Shang, X.~Wang, Y.-S. Liu, and M.~Zwicker, ``Y2seq2seq: Cross-modal
  representation learning for 3d shape and text by joint reconstruction and
  prediction of view and word sequences,'' in \emph{Proceedings of the AAAI
  Conference on Artificial Intelligence}, vol.~33, no.~01, 2019, pp. 126--133.

\bibitem{text2mesh}
O.~Michel, R.~Bar-On, R.~Liu, S.~Benaim, and R.~Hanocka, ``Text2mesh:
  Text-driven neural stylization for meshes,'' in \emph{Proceedings of the
  IEEE/CVF Conference on Computer Vision and Pattern Recognition}, 2022, pp.
  13\,492--13\,502.

\bibitem{clip-forge}
A.~Sanghi, H.~Chu, J.~G. Lambourne, Y.~Wang, C.-Y. Cheng, M.~Fumero, and K.~R.
  Malekshan, ``Clip-forge: Towards zero-shot text-to-shape generation,'' in
  \emph{Proceedings of the IEEE/CVF Conference on Computer Vision and Pattern
  Recognition}, 2022, pp. 18\,603--18\,613.

\bibitem{clip}
A.~Radford, J.~W. Kim, C.~Hallacy, A.~Ramesh, G.~Goh, S.~Agarwal, G.~Sastry,
  A.~Askell, P.~Mishkin, J.~Clark \emph{et~al.}, ``Learning transferable visual
  models from natural language supervision,'' in \emph{International Conference
  on Machine Learning}.\hskip 1em plus 0.5em minus 0.4em\relax PMLR, 2021, pp.
  8748--8763.

\bibitem{3Drec_mem}
S.~Yang, M.~Xu, H.~Xie, S.~Perry, and J.~Xia, ``Single-view 3d object
  reconstruction from shape priors in memory,'' in \emph{Proceedings of the
  IEEE/CVF Conference on Computer Vision and Pattern Recognition}, 2021, pp.
  3152--3161.

\bibitem{transformer}
A.~Vaswani, N.~Shazeer, N.~Parmar, J.~Uszkoreit, L.~Jones, A.~N. Gomez,
  {\L}.~Kaiser, and I.~Polosukhin, ``Attention is all you need,''
  \emph{Advances in neural information processing systems}, vol.~30, 2017.

\bibitem{bert}
J.~Devlin, M.-W. Chang, K.~Lee, and K.~Toutanova, ``Bert: Pre-training of deep
  bidirectional transformers for language understanding,'' \emph{arXiv preprint
  arXiv:1810.04805}, 2018.

\bibitem{convirt}
Y.~Zhang, H.~Jiang, Y.~Miura, C.~D. Manning, and C.~P. Langlotz, ``Contrastive
  learning of medical visual representations from paired images and text,''
  \emph{arXiv preprint arXiv:2010.00747}, 2020.

\bibitem{pke}
F.~Boudin, ``Pke: an open source python-based keyphrase extraction toolkit,''
  in \emph{Proceedings of COLING 2016, the 26th international conference on
  computational linguistics: system demonstrations}, 2016, pp. 69--73.

\bibitem{yue2020interventional}
Z.~Yue, H.~Zhang, Q.~Sun, and X.-S. Hua, ``Interventional few-shot learning,''
  \emph{Advances in neural information processing systems}, vol.~33, pp.
  2734--2746, 2020.

\bibitem{adam}
D.~P. Kingma and J.~Ba, ``Adam: A method for stochastic optimization,''
  \emph{arXiv preprint arXiv:1412.6980}, 2014.

\bibitem{imle}
K.~Li and J.~Malik, ``Implicit maximum likelihood estimation,'' \emph{arXiv
  preprint arXiv:1809.09087}, 2018.

\bibitem{shapenet}
A.~X. Chang, T.~Funkhouser, L.~Guibas, P.~Hanrahan, Q.~Huang, Z.~Li,
  S.~Savarese, M.~Savva, S.~Song, H.~Su \emph{et~al.}, ``Shapenet: An
  information-rich 3d model repository,'' \emph{arXiv preprint
  arXiv:1512.03012}, 2015.

\bibitem{inception}
C.~Szegedy, V.~Vanhoucke, S.~Ioffe, J.~Shlens, and Z.~Wojna, ``Rethinking the
  inception architecture for computer vision,'' in \emph{Proceedings of the
  IEEE conference on computer vision and pattern recognition}, 2016, pp.
  2818--2826.

\end{thebibliography}

\ifCLASSOPTIONcaptionsoff
  \newpage
\fi



%
\end{document}